\documentclass[runningheads]{llncs}
\usepackage{graphicx}

\usepackage{mathtools,tikz}
\usetikzlibrary{tikzmark,decorations.pathreplacing}
\usepackage{comment}
\usepackage{color}

\usepackage{graphicx}
\usepackage{amsmath}
\usepackage{amssymb}
\usepackage{booktabs}

\usepackage[pagebackref,breaklinks,colorlinks]{hyperref}

\usepackage[capitalize]{cleveref}
\crefname{section}{Sec.}{Secs.}
\Crefname{section}{Section}{Sections}
\Crefname{table}{Table}{Tables}
\crefname{table}{Tab.}{Tabs.}

\usepackage[utf8]{inputenc} 
\usepackage[T1]{fontenc}    
\usepackage{url}            
\usepackage{amsfonts}       
\usepackage{nicefrac}       
\usepackage{microtype}      
\usepackage{listings}
\usepackage{wrapfig}
\usepackage{floatflt}

\usepackage{multirow}
\usepackage{graphicx}
\usepackage{float}
\usepackage{algorithm,algorithmicx,algpseudocode}
\usepackage{dblfloatfix}
\usepackage{makecell}
\usepackage{orcidlink} 
\usepackage[misc]{ifsym}

\usepackage{xcolor}  
\usepackage{colortbl}  
\definecolor{gray}{rgb}{0.5,0.5,0.5}
\definecolor{darkergreen}{RGB}{34, 139, 34}
\definecolor{gray94}{gray}{.94}

\newcommand{\green}[1]{\textcolor{darkergreen}{#1}}
\newcommand{\blue}[1]{\textcolor{blue}{#1}}
\newcommand{\gray}[1]{\textcolor{gray}{#1}}

\newcommand{\fn}[1]{\footnotesize{#1}}

\newcommand{\gbf}[1]{\green{\bf{#1}}}
\newcommand{\gfn}[1]{\green{\fn{#1}}}

\usepackage{pifont}%
\usepackage{xspace}
\usepackage{placeins}
\newcommand{\cmark}{\ding{51}\xspace}%
\newcommand{\xmarkg}{\textcolor{gray}{\ding{55}}\xspace}%

\begin{document}
\pagestyle{headings}
\mainmatter
\def\ECCVSubNumber{3311}  

\title{AutoMix: Unveiling the Power of Mixup for Stronger Classifiers} 

\titlerunning{AutoMix}
%
\author{
    Zicheng Liu\inst{1,2\star}\orcidlink{0000-0003-1106-2963} \and
    Siyuan Li\inst{1,2\star}\orcidlink{0000-0001-6806-2468} \and
    Di Wu\inst{1,2}\and
    Zihan Liu\inst{1,2}\and
    Zhiyuan Chen\inst{2}\and
    Lirong Wu\inst{1,2}\and
    Stan Z. Li\inst{2,}${^{\textrm{\Letter}}}$\orcidlink{0000-0002-2961-8096}
}
%
\authorrunning{Liu. et al.}
%
\institute{
Zhejiang University, Hangzhou, 310000, China \and
AI Lab, School of Engineering, Westlake University, Hangzhou, 310000, China \\
\email{\{liuzicheng, lisiyuan, wudi, liuzihan, chenzhiyuan, wulirong, stan.z.li\}@westlake.edu.cn}\\
$\star$ Equal contribution, \textrm{\Letter} Corresponding author
}
\maketitle

\begin{abstract}
Data mixing augmentation have proved to be effective in improving the generalization ability of deep neural networks. 
While early methods mix samples by hand-crafted policies (\textit{e.g.}, linear interpolation), recent methods utilize saliency information to match the mixed samples and labels via complex offline optimization. 
However, there arises a trade-off between precise mixing policies and optimization complexity. To address this challenge, we propose a novel automatic mixup (AutoMix) framework, where the mixup policy is parameterized and serves the ultimate classification goal directly. 
Specifically, AutoMix reformulates the mixup classification into two sub-tasks (\textit{i.e.}, mixed sample generation and mixup classification) with corresponding sub-networks and solves them in a bi-level optimization framework.
For the generation, a learnable lightweight mixup generator, Mix Block, is designed to generate mixed samples by modeling patch-wise relationships under the direct supervision of the corresponding mixed labels.
To prevent the degradation and instability of bi-level optimization, we further introduce a momentum pipeline to train AutoMix in an end-to-end manner. 
Extensive experiments on nine image benchmarks prove the superiority of AutoMix compared with state-of-the-art in various classification scenarios and downstream tasks.

\end{abstract}

\section{Introduction} \label{sec:intro}
Recent years have witnessed the great success of Deep Neural Networks (DNNs) in various tasks, such as image processing~\cite{2022dlme,2021genurl,wu2022deepclustering,tan2022HCR,tan2021colearning}, graph learning~\cite{xia2022pre,wu2021self,cheng2022physical}, and video processing~\cite{ijcai2020tlpg,liu2021sg,cui2021tf,liu2021densernet,zhao2022tracking}. 
Most of these successes can be attributed to the use of complex network architectures with numerous parameters and a sufficient amount of data. 
However, when the data is insufficient, models with high complexity, \textit{e.g.}, Transformer-based networks~\cite{iclr2021vit,icml2021deit}, are prone to over-fitting and overconfidence~\cite{guo2017calibration}, resulting in poor generalization abilities~\cite{wan2013regularization,srivastava2014dropout,bishop2006pattern}.

\begin{figure}
    \centering
    \vspace{-0.5em}
    \includegraphics[width=1.0\linewidth]{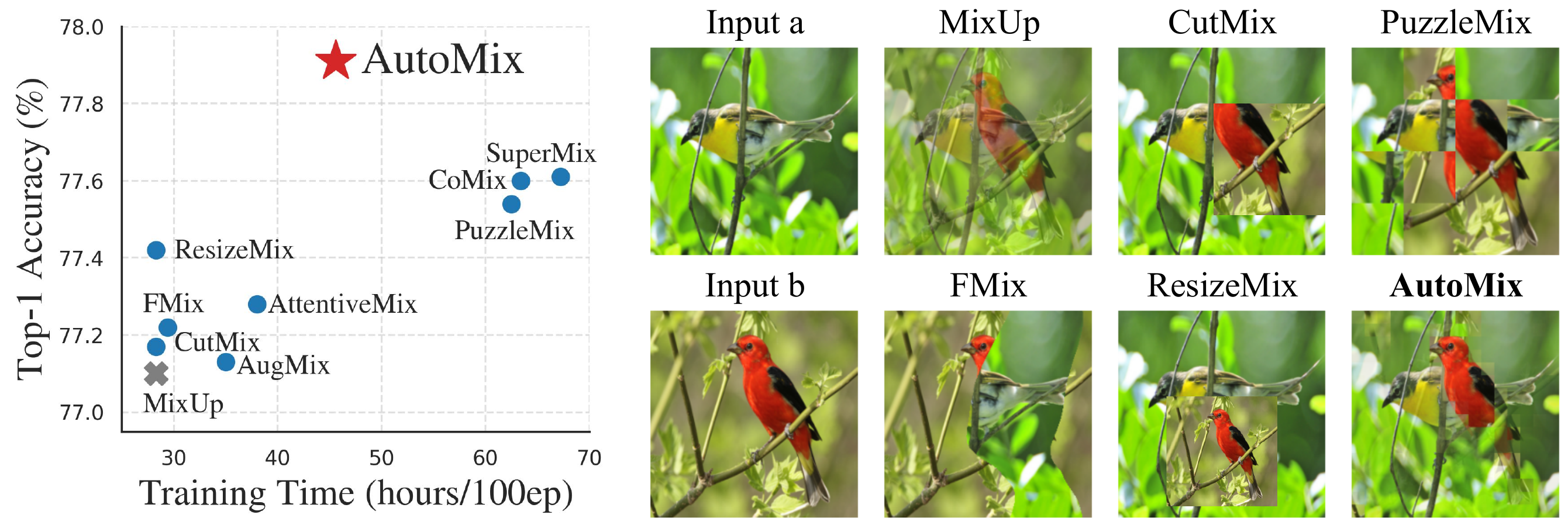}
    \vspace{-8mm}
    \caption{
    The plot of efficiency \textit{vs.} accuracy on ImageNet-1k and visualization of mixup methods. AutoMix improves performance without the heavy computational overhead.}
    \vspace{-2.5em}
    \label{fig:intro}
\end{figure}

To improve the generalization of DNNs, a series of data mixing augmentation techniques emerged. As shown in Figure~\ref{fig:intro}, MixUp~\cite{zhang2017mixup} generates augmented samples via a linear combination of corresponding data pairs; CutMix~\cite{yun2019cutmix} designs a patch replacement strategy that randomly replaces a patch in an image with patches from the other image. However, these \textit{hand-crafted} methods~\cite{verma2019manifold,faramarzi2020patchup,harris2020fmix} cannot guarantee mixed samples containing target objects and might cause the \textit{label mismatch} problem. 
Subsequently, \cite{icassp2020attentive,uddin2020saliencymix,qin2020resizemix} try to guide CutMix by saliency information to relieve this problem. 
Recently, \textit{optimization-based} methods tried to solve the problem by searching an approximate mixing policy~\cite{dabouei2021supermix,kim2020puzzle,kim2021comixup} based on portfolio optimization, \textit{e.g.}, maximizing the saliency regions to confirm the co-presence of the targets in the mixed samples. 
Although they design more precise mixing policies than \textit{hand-crafted} methods, their indirect optimization and heavy computational overhead limit the algorithms' efficiency. Evidently, it is not efficient to transform the mixup policy from a random linear interpolation to a complex portfolio optimization problem. 

This paper mainly discusses two questions: \textbf{(1) how to design an accurate mixing policy and serve directly to the mixup classification objective; (2) how to solve generation-classification optimization problems efficiently instead of portfolio optimizations.}
As a basis for solving these two issues, we first reformulate the mixup training into two sub-tasks, mixed sample generation and mixup classification. 
Then, we propose a novel automatic mixup framework (AutoMix) that generates accurate mixed samples by a generation sub-network, Mix Block (MB), with a good complexity-accuracy trade-off. 
Specifically, MB is a cross-attention-based module that dynamically selects discriminative pixels based on feature maps of the sample pair to match the corresponding mixed labels. 
However, MB may collapse into trivial solutions when optimized jointly with the classification encoder due to a gradient entanglement problem. 
Thus, Momentum Pipeline (MP) is further introduced to stabilize AutoMix and decouple the training process of this bi-level optimization problem.
Comprehensive experiments on eight classification benchmarks (CIFAR-10/100, Tiny-ImageNet, ImageNet-1k, CUB-200, FGVC-Aircraft, iNaturalist2017/2018, and Places205) and eight network architectures show that AutoMix consistently outperforms state-of-the-art mixup methods across different tasks. 
We further provide extensive analysis to verify the effectiveness of proposed components and the robustness of hyper-parameters. 
Our main contributions are three-fold:
\begin{itemize}
    \item From a fresh perspective, we divide the mixup training into bi-level subtasks: mixed sample generation and mixup classification, and regard the generation as an auxiliary task to the classification. We unify them into a framework named AutoMix to optimize the mixup policy in an end-to-end manner.
    \item A novel Mix Block is designed for mixed sample generation. The combination of Mix Block and Momentum Pipeline optimizes the two sub-tasks in a decoupled manner and improves mixup training accuracy and stability.
    \item AutoMix surpasses counterparts significantly on various classification scenarios based on eight popular network architectures and downstream tasks.
\end{itemize}

\vspace{-1.0em}
\section{Preliminaries}
\label{sec:preliminaries}
\vspace{-0.25em}
\paragraph{\textbf{Mixup training.}} We first consider the general image classification task with $k$ different classes: given a finite set of $n$ samples $X=[x_i]_{i=1}^{n}\in \mathbb{R}^{n\times W\times H\times C}$ and their ground-truth class labels $Y = [y_i]_{i=1}^{n}\in \mathbb{R}^{n\times k}$, encoded by a one-hot vector $y_i\in \mathbb{R}^{k}$. 
We seek the mapping from the data $x_i$ to its class label $y_i$ modeled by a deep neural network $f_{\theta}:x\longmapsto y$ with network parameters $\theta$ by optimizing a classification loss $\ell(.)$, say the cross entropy (CE) loss,
\begin{equation}
    \vspace{-1pt}
	\ell_{CE}(f_{\theta}(x), y) = -y\log f_{\theta}(x).
    \vspace{-1pt}
    \label{eq:basic_cls}
\end{equation}
Then we consider the mixup classification task: given a sample mixup function $h$, a label mixup function $g$, and a mixing ratio $\lambda$ sampled from $Beta(\alpha,\alpha)$ distribution, we can generate the mixup data $X_{mix}$ with $x_{mix}=h(x_i,x_j,\lambda)$ and the mixup label $Y_{mix}$ with $y_{mix}=g(y_i,y_j,\lambda)$. Similarly, we learn $f_{\theta}:x_{mix}\longmapsto y_{mix}$ by mixup cross-entropy (MCE) loss,
\begin{equation}
    \vspace{-1pt}
	 \ell_{MCE} = \lambda \ell_{CE}(f_\theta(x_{mix}), y_i) +
	              (1-\lambda) \ell_{CE}(f_\theta(x_{mix}), y_j).
	\vspace{-1pt}
    \label{eq:mixup_cls}
\end{equation}

\paragraph{\textbf{Mixup reformulation.}}
Comparing Eq.~\ref{eq:basic_cls} and Eq.~\ref{eq:mixup_cls}, the mixup training has the following features: 
(1) extra mixup policies, $g$ and $h$, are required to generate $X_{mix}$ and $Y_{mix}$. 
(2) the classification performance of $f_{\theta}$ depends on the generation policy of mixup. Naturally, we can split the mixup task into two complementary sub-tasks: 
(i) mixed sample generation and (ii) mixup classification. 
Notice that the sub-task (i) is subordinate to (ii) because the final goal is to obtain a stronger classifier. 
Therefore, from this perspective, we regard the mixup generation as an auxiliary task for the classification task. 
Since $g$ is generally designed as a linear interpolation, i.e., $g(y_i,y_j,\lambda) = \lambda y_i + (1-\lambda)y_j$, $h$ becomes the key function to determine the performance of the model. 
Generalizing previous offline methods, we define a parametric mixup policy $h_{\phi}$ as the sub-task with another set of parameters $\phi$. 
The final goal is to optimize $\ell_{MCE}$ given $\theta$ and $\phi$ as below:
\begin{equation}
    \vspace{-1pt}
	\mathop{{\rm min}}\limits_{\theta,~\phi} \ell_{MCE} \Big (f_{\theta} \big (h_{\phi}(x_i,x_j,\lambda) \big), g(y_i,y_j,\lambda) \Big).
	\vspace{-1pt}
    \label{eq:emix_cls}
\end{equation}

\begin{figure}[t]
    \centering
    \vspace{-0.5em}
    \includegraphics[width=0.95\linewidth]{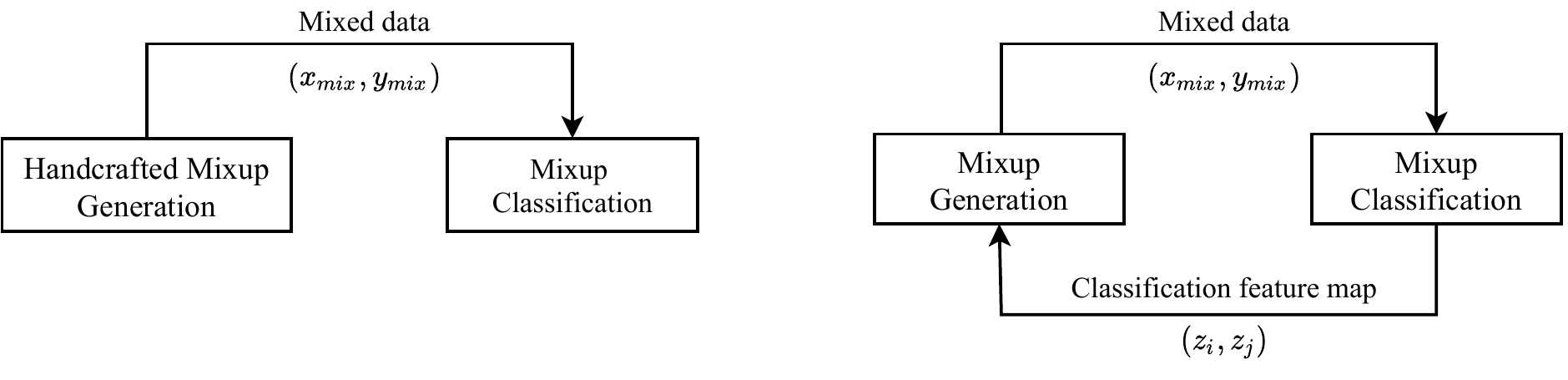}
    \vspace{-14pt}
    \caption{The difference between AutoMix and offline approaches. \textbf{Left}: Offline mixup methods, where a fixed mixup policy generates mixed samples for the classifier to learn from. \textbf{Right}: AutoMix, where the mixup policy is trained with the feature map.}
    \label{fig:semantic}
    \vspace{-1.5em}
\end{figure}

\paragraph{\textbf{Offline mixup limits the power of mixup.}}
Keep the reformulation in mind, the previous methods focus on manually designing $h(\cdot)$ in an offline and non-parametric manner based on their prior hypotheses, or arguably, such mixup policies are separated from the ultimate optimization of the model, e.g., an optimization algorithm with the goal of maximizing saliency information.
Specifically, they build an implicit connection between the two sub-tasks, as shown on the left of Figure~\ref{fig:semantic}. 
Therefore, the mixed samples generated from these offline mixup policies could be redundant or mislead the training. 
To address this, we propose AutoMix, \textit{which combines these two sub-tasks in a mutually beneficial manner and unveils the power of mixup.}

\section{AutoMix}
\begin{wrapfigure}{r}{0.5\textwidth}
    \centering
    \vspace{-68pt}
    \includegraphics[width=1.0\linewidth]{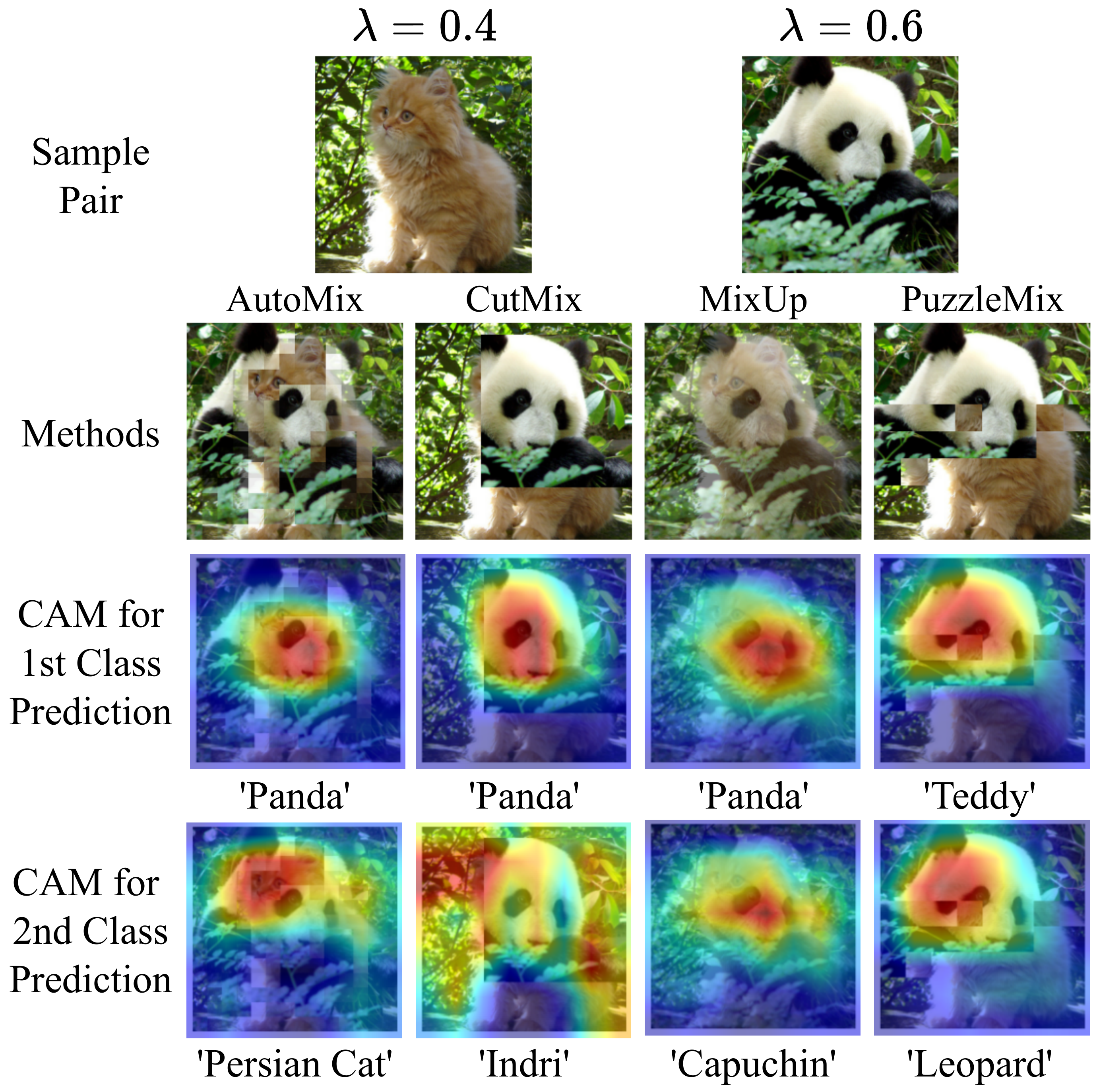}
    \vspace{-19pt}
    \caption{Illustration of \textit{label mismatch} by visualizing mixed samples and class activation mapping (CAM)~\cite{selvaraju2017gradcam} on `Panda' and `Persian Cat'. 
    From top to bottom rows, we show the original images, mixed images, and CAM for top-2 predicted classes, respectively. 
    }
    \vspace{-43pt}
    \label{fig:CAM}
\end{wrapfigure}
We build a bridge between the mixup generation and classification task with a unified optimization framework named as AutoMix to improve the mixup training efficiency. In this framework, the proposed Mix Block (MB) and Momentum Pipeline (MP) in AutoMix not only can generate semantic mixed samples but reduces computational overhead significantly.
A comparison overview with offline approaches is presented in Figure~\ref{fig:semantic}. 

\subsection{Label Mismatch: MixBlock}
\begin{figure}[t]
    \centering
    \includegraphics[width=1.0\linewidth]{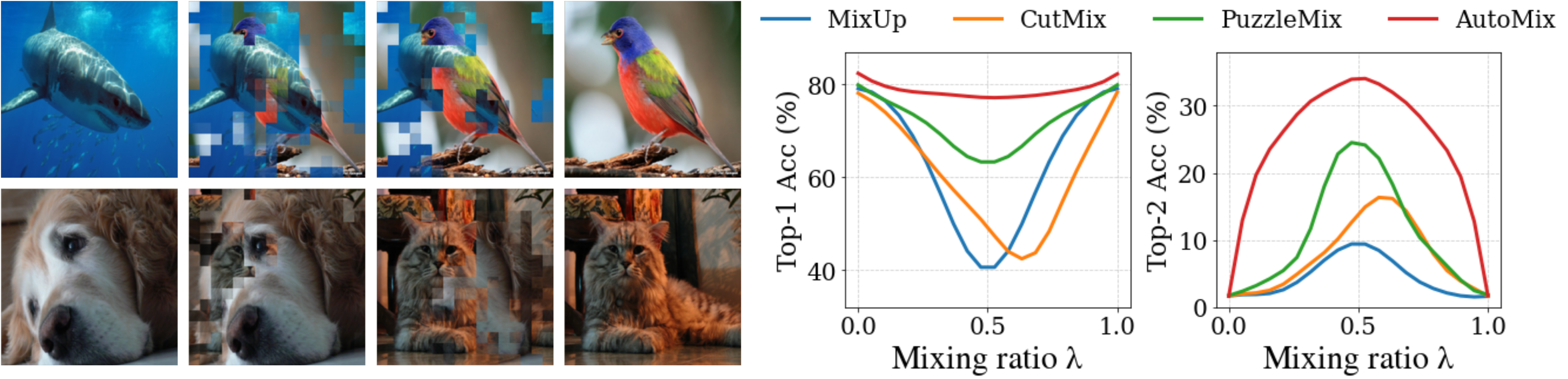}
    \vspace{-18pt}
    \caption{\textbf{Left}: AutoMix samples with different $\lambda$ (0, 0.3, 0.7, 1).
    \textbf{Right}: Top-1 accuracy of mixed data. Prediction is counted as correct if the top-1 prediction belongs to $\{y_i, y_j\}$; Top-2 accuracy is calculated by counting the top-2 predictions are equal to $\{y_i, y_j\}$.}
    \label{fig:lambda_main}
    \vspace{-14pt}
\end{figure}

In Figure~\ref{fig:CAM}, we further examined that offline approaches are incapable of addressing the \textit{label mismatch} issue in mixup training.
It is difficult for offline methods to preserve the discriminative features in the mixed sample if detached from the final optimization goal.
As a result, the prediction of the accuracy of the mixed sample is limited (see the right of Figure \ref{fig:lambda_main}).
This paper presents a parametric mixup generation function named Mix Block (MB) $\mathcal{M}_{\phi}$ for learning a mixup policy without requiring extensive saliency computation.
$\mathcal{M}_\phi$ generates a pixel-wise mixup mask $s\in \mathbb{R}^{H\times W}$ for the pairs of input images, where $s_{w,h}\in [0,1]$. 
We regard the mask-based mixup policy as an adaptive selection process in terms of $\lambda$, which can automatically select the discriminative patches from sample pairs to generate label-matched mixed samples. 
Thus, the core of $\mathcal{M}_\phi$ is the devised $\lambda$ embedded cross-attention mechanism to learn the pixel-level proportional relationships in a given data pair. 
To do so, the deep feature maps $z$ from $f_{\theta}$ with rich spatial and semantic information can be utilized to \textit{bootstrap the two sub-tasks of mixup}. Additionally, to facilitate the capture of task-relevant information in the generated mixed samples, the $\mathcal{M}_\phi$ training is directly supervised by the target loss, $\ell_{MCE}$, in an end-to-end manner.

\begin{figure}[t]
    \centering
    \includegraphics[width=1.0\linewidth]{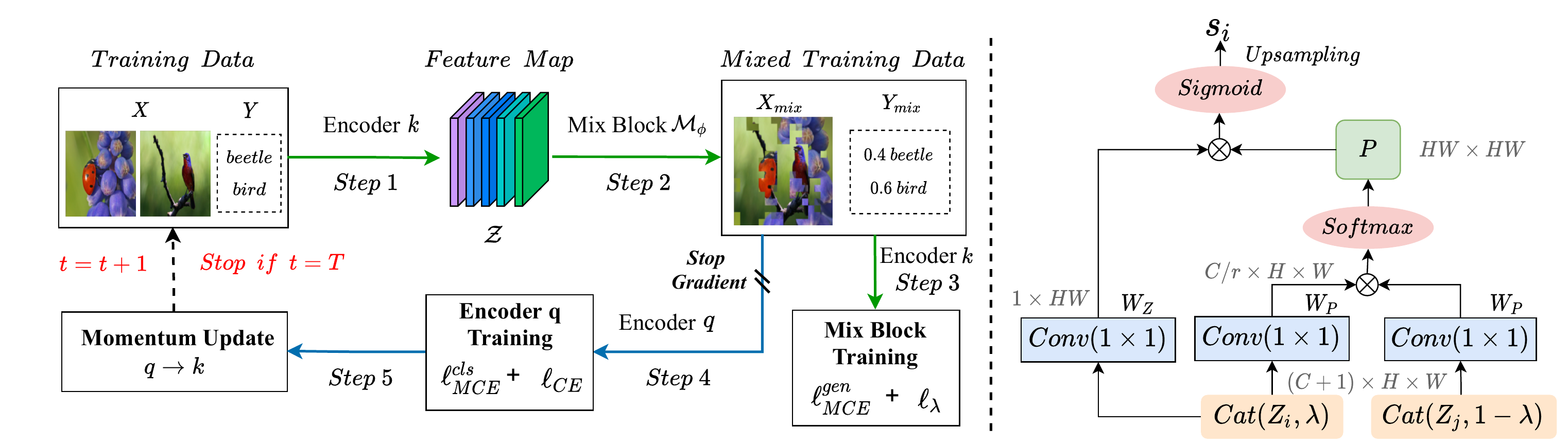}
    \vspace{-2.0em}
    \caption{
    The \textbf{left} diagram represents the five key steps of AutoMix. 
    (1) Extract feature map $\mathcal{Z}$ from the frozen encoder $k$. 
    (2) Mix Block $\mathcal{M}_\phi$ generates mixed samples by using $\mathcal{Z}$ and mixup ratio $\lambda\in [0, 1]$. 
    (3) and (4) Decoupled training $\mathcal{M}_\phi$ and encoder $q$ via \textit{stop gradient}, the blue and green lines indicate the encoder training and the $\mathcal{M}_\phi$ training, correspondingly. 
    (5) Update the $k$'s parameters through momentum moving. The \textbf{right} diagram is the architecture of proposed $\mathcal{M}_\phi$. 
    }
    \label{fig:framework}
    \vspace{-1.25em}
\end{figure}

\vspace{-8pt}
\paragraph{\textbf{Parametric mixup generation.}}
The generation task can be formulated as a dynamic regression problem: given a sample pair $(x_i, x_j)$ and a mixing ratio $\lambda$, MB predicts the probability that each pixel (or patch) on $x_{mix}$ belongs to $x_i$ according to the feature map pair $(z_i, z_j)$ and mixing ratio $\lambda$. 
The overall parametric mixup function of AutoMix can be formulated as follows:
\begin{equation}
    \vspace{-1pt}
    \begin{split}
        h_{\phi}(x_i, x_j, \lambda) = 
        \mathcal{M_{\phi}}(z_{i,\lambda}^l, z_{j, 1-\lambda}^l) \odot x_i
        + (1-\mathcal{M_{\phi}}(z_{i, \lambda}^l, z_{j, 1-\lambda}^l)) \odot x_j,
    \end{split}
    \label{eq:mix_function}
    \vspace{-1pt}
\end{equation}
where $\odot$ denotes element-wise product; $z^l_\lambda$ is $\lambda$ embedded feature map at $l$-th layer. 
As shown in the right of Figure \ref{fig:framework}, we first embed $\lambda$ with the $l$-th feature map in a simple and efficient way by concatenating, $z^{l}_{\lambda} = {\rm concat}(z, \lambda)$, whose effectiveness has been shown in the left of Figure~\ref{fig:lambda_main}.
As we can see from Equation~\ref{eq:mix_function}, our aim is to obtain a pixel-level mask $s$ in the input space from $\mathcal{M}_\phi(\cdot)$ based on $\lambda$ embedded $z_{i,\lambda}^l$ and $z_{j,1-\lambda}^l$ to generate semantic mixed samples.
In order to achieve this goal, a pair-wise similarity matrix $P$ and an upsampling function $U(\cdot)$ is required.
Due to the symmetry of mixup, i.e., the sum of the two masks used to generate a mixed sample is equal to 1, for $x_i$ of a pair $(x_i, x_j)$, we can denote $\mathcal{M}_{\phi}:z_{i,\lambda}^l,z_{j,1-\lambda}^l\longrightarrow s_i$, 
\begin{equation}
    \vspace{-2pt}
    s_{i} = U \Big (\sigma \big (P(z^l_{i, \lambda},z^l_{j, 1-\lambda}) \otimes W_{Z}\, z^l_{i, \lambda}\big ) \Big ),
    \vspace{-2pt}
    \label{eq:mixblock_score}
\end{equation}
where $W_{Z}$ is a linear transformation matrix; $\sigma$ is the Sigmoid activation function, which is used to probabilize the mask; and $s_{i}$ is the $H\times W$ mask we are looking for.
By multiplying $P$ and the value embedding, $W_z z^l_{i,\lambda}$, the discriminative features in $x_{i,\lambda}$ relative to $x_{j,1-\lambda}$ are then selected.
Symmetrically, the mask $s_j$ for $x_j$ can be calculated in this way, $s_j=1-s_i$.
Furthermore, the similarity matrix $P$ has to consider both $\lambda$ information and relative relationships in a sample pair; thus, the \textit{cross-attention mechanism} is introduced to achieve this purpose. 
When $x_i$ in a sample pair $(x_i, x_j)$ is taken as the input, a mask can be generated dynamically from corresponding $z_{i,\lambda}^l$ and $P$ matrix. 
Formally, our cross-attention can be formulated as:
\begin{equation}
    \vspace{-2pt}
    P(z^l_{i,\lambda},z^l_{j,1-\lambda}) =
    {\rm{softmax}} \Big (\frac{(W_{P}\,z^l_{i,\lambda})^{T} \otimes  W_{P}\,z^l_{j,1-\lambda}}{C(z^l_{i,\lambda},z^l_{j,1-\lambda})} \Big ),
    \vspace{-2pt}
    \label{eq:pairwise_weight}
\end{equation}
where $W_{P}$ denotes shared linear transformation matrices (e.g., 1$\times$1 convolution), $\otimes$ denotes matrix multiplication, and $C(z^l_{i,\lambda},z^l_{j,1-\lambda})$ is a normalization factor. 
Notice that $P$ is the row normalized pair-wise similarity matrix between every spatial position on $z^l_{i,\lambda}$ and $z^l_{j,1-\lambda}$.
Similarly, if we take $z_{j,1-\lambda}^l$ as the value, then the mask can be computed by transposing $P$ and $s_i=1-s_j$. 

\vspace{-4pt}
\paragraph{\textbf{AutoMix in end-to-end training.}}
The framework is shown in Figure~\ref{fig:framework}, given a set of labeled data $\mathcal{D}=\{(x_i, y_i)\}_{i=1}^{n}$ and the corresponding $l$-th layer feature map $\mathcal{Z} = \{z_i^l\}_{i=1}^{n}$, $\mathcal{M}_\phi$ is nested in encoder for optimization. 
Under the supervision of the same loss $\ell_{MCE}$, the encoder is trained using the mixed sample generated by $\mathcal{M}_\phi$, which in turn uses the backbone's feature to generate the mixed sample.
To enable $\mathcal{M}_\phi$ to find the $\lambda$ correspondence between the $x_{mix}$ and $y_{mix}$ at the early stage of training, our auxiliary loss is proposed:
\begin{equation}
    \vspace{-2pt}
    \ell_{\lambda}=\gamma \max \big (\vert|\lambda - \frac{1}{HW}\sum_{h,w}s_{i,h,w}\vert| - \epsilon, 0 \big),
    \vspace{-2pt}
    \label{eq:mask_loss}
\end{equation}
where $\gamma$ is a loss weight linearly decreased to $0$ during training. We set the initial $\gamma$ to $0.1$ and $\epsilon=0.1$. 
Notice that AutoMix uses standard cross-entropy loss $\ell_{CE}$ as default. 
$\ell_{CE}$ loss facilitates the backbone to provide a stable feature map at the early stage so that speeds up $\mathcal{M}_\phi$ converges. 
To differentiate the function of $\ell_{MCE}$, $cls$ denotes classification task for training encoder and $gen$ denotes generation task for training $\mathcal{M}_\phi$. AutoMix can be optimized by a joint loss:
\begin{equation}
    \vspace{-1pt}
    \mathcal{L}(\theta, \phi) = \underbrace{\ell_{CE} + 
    \ell_{MCE}^{cls}}_{classification} +
    \underbrace{\ell_{MCE}^{gen} + 
    \ell_{\lambda}}_{generation}.
    \label{eq:loss}
    \vspace{-1pt}
\end{equation}
Obviously, the purpose of the classification task is to optimize $\theta$ while the generation task is to optimize $\phi$. 
Therefore, this is a typical bi-level optimization problem.
Although $\mathcal{M}_\phi$ does not need extra computational overhead to maximize the saliency information, using SGD to directly update the nested $\theta$ and $\phi$ will lead to instability. 
To address this problem properly, we use the momentum pipeline to decouple the training of $\theta$ and $\phi$. 
As indicated in Eq.~\ref{eq:loss}, though the same $\ell_{MCE}$ is used, the focus of each is different.

\vspace{-5pt}
\subsection{Bi-level Optimization: Momentum Pipeline}
\label{subsec:MP}
\begin{wrapfigure}{r}{0.5\linewidth}
    \centering
    \vspace{-24pt}
    \includegraphics[width=1.0\linewidth]{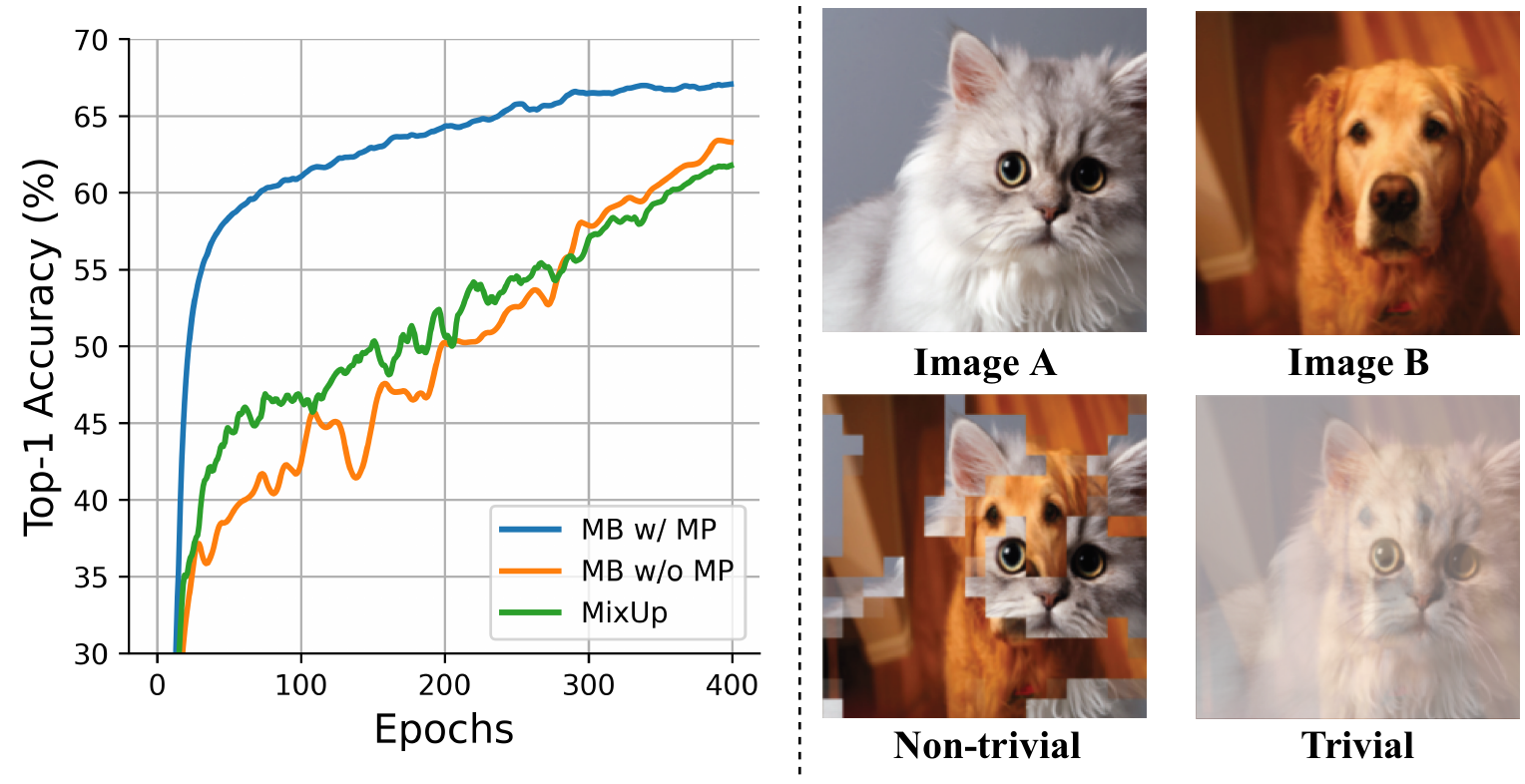}
    \vspace{-18pt}
    \caption{Accuracy on Tiny-ImageNet and different results of the mixed sample. Momentum pipeline decoupled mixup generation and classification, which mitigates the trivial solution problem.}
    \label{fig:collapse}
    \vspace{-18pt}
\end{wrapfigure}
\begin{figure*}[t]
    \centering
    \includegraphics[width=1.0\linewidth]{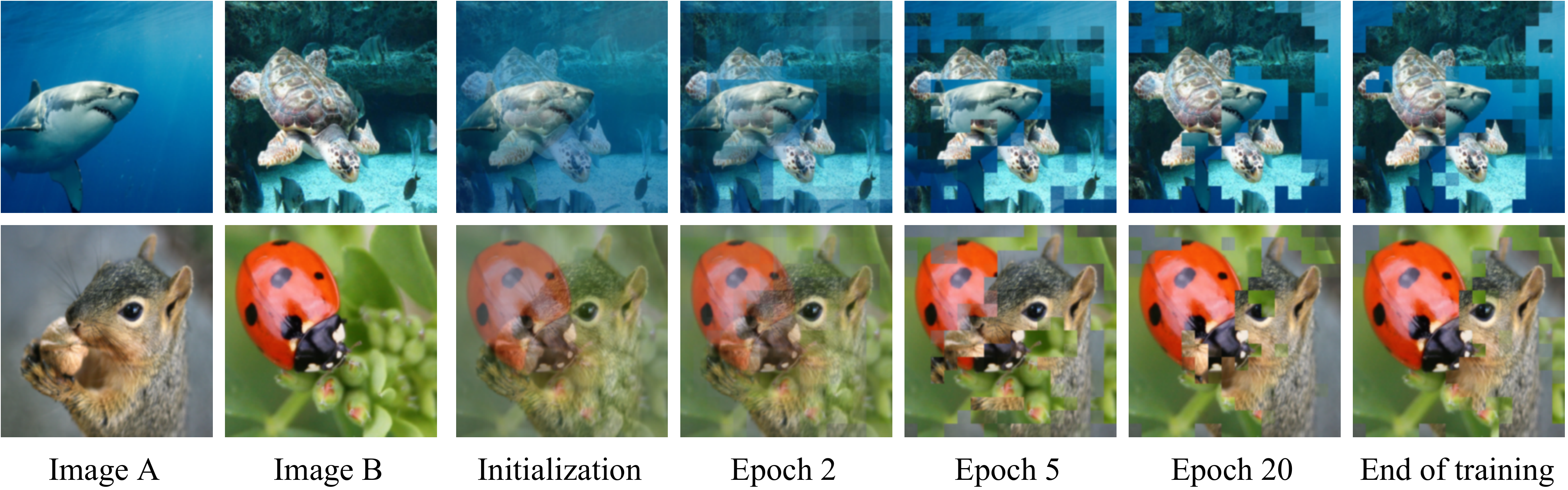}
    \vspace{-18pt}
    \caption{Visualization of mixed samples generated by $\mathcal{M}_\phi$ with $\lambda=0.5$ at different training periods on ImageNet-1k (100 epochs in total). It is worth noting that $\mathcal{M}_\phi$ is able to generate mixed samples stably and converge quickly with the addition of MP.}
    \label{fig:convergence}
    \vspace{-15pt}
\end{figure*}

Although MB is designed to be lightweight and efficient, it also poses a bi-level optimization problem with \textit{gradient entanglement}.
Experiments demonstrate that the entanglement problem may cause $\mathcal{M}_\phi$ trapped into a trivial solution (degraded to MixUp, in Figure~\ref{fig:collapse}).
$\mathcal{M}_\phi$ with much smaller parameters than the encoder will be disturbed by the classification task when optimizing both the two sub-tasks at the same time. MB thus cannot generate semantic mixed samples stably and eventually collapse.
According to Eq.~\ref{eq:emix_cls} and Eq.~\ref{eq:loss}, for each iteration, the gradient entanglement problem of $\mathcal{L}^{cls}$ in $\mathcal{M}_\phi$ can be formulated as
\begin{equation}
    \vspace{-1pt}
    \nabla_{\phi}\mathcal{L}_{MCE}^{cls} \propto \nabla_{\phi}h_{\phi}(x_i, x_j, \lambda) \odot f'_{\theta}(h_{\phi}(x_i, x_j, \lambda)).
    \vspace{-1pt}
\end{equation}
It is notable that the instability of $f_\theta$ may result in a vicious cycle of joint training. 
As a consequence, the primary goal of getting the Eq.~\ref{eq:emix_cls} operating well is to ensure that $f_\theta$ outputs stable features and, to the extent possible, that $\phi$ and $\theta$ can focus on their own tasks in the case of using the same loss.
Inspired by methods in self-supervised learning~\cite{he2020momentum,nips2020byol}, they adopted momentum pipeline (MP) to avoid the feature collapse and realized that the teacher network $f_{\theta_k}$ of the Siamese network shows more stable performance than student network $f_{\theta_q}$. 
Along this path, we designed a new MP for decoupling the nested bi-level optimization problem of AutoMix: the student network $f_{\theta_q}$ focuses on the classification task, while the stable teacher network $f_{\theta_k}$ is connected with $\mathcal{M}_\phi$ to perform generation task.
Moreover, optimizing Eq.~\ref{eq:loss} with batch approach requires $X_{mix}$ generated by $f_{\theta_k}$ and $\mathcal{M}_\phi$ first and then using $X_{mix}$ to optimize $f_{\theta_q}$.
By analogy, referring to the Expectation-Maximization (EM) algorithm, the two sets of parameters $\theta$ and $\phi$ can be optimized in an alternating way by the designed MP, i.e., first fix one set of parameters optimizing the other:
\begin{align}
    \vspace{-2pt}
    \theta_{q}^{t} &\leftarrow \mathop{{\rm argmin}}\limits_\theta \mathcal{L}(\theta_{q}^{t-1}, \phi^{t-1}), \label{eq:qem} \\
    \phi^{t} &\leftarrow \mathop{{\rm argmin}}\limits_\phi \mathcal{L}(\theta_{k}^{t}, \phi^{t-1}), \label{eq:kem}
    \vspace{-2pt}
\end{align}
where $t$ is the iteration step, $\theta_q$ and $\theta_k$ represent the parameters of the student and teacher network, respectively.
Note that $f_{\theta_q}$ and $f_{\theta_k}$ share the same network structure with the same initialized parameters, but $f_{\theta_k}$ is updated via an exponential moving average (EMA) strategy~\cite{siam1992ema} from $f_{\theta_q}$:
\begin{equation}
    \vspace{-2pt}
    \theta_k\leftarrow m\theta_k+(1-m)\theta_q,
    \label{eq:momentum}
    \vspace{-2pt}
\end{equation}
where $m\in [0,1)$ is the momentum coefficient. 
It is worthy to notice that \textit{MP not only solves optimization instability but also significantly speeds up and stabilizes the convergence of AutoMix.} 
In Figure~\ref{fig:convergence}, $\mathcal{M}_\phi$ gets close to convergence in the first few epochs and consistently delivers high-quality mixed samples to $f_{\theta}$.
Moreover, detailed AutoMix architecture and pseudo code are provided in Appendix.
\section{Experiments}
\label{sec:expt}
We evaluate AutoMix in three aspects: 
(1) Image classification in various scenarios based on various network architectures, 
(2) Robustness against corruptions and adversarial samples, and 
(3) Transfer learning capacities to downstream tasks. 

\vspace{-8pt}
\subsection{Evaluation on Image Classification}
\label{exp:cls}
This subsection demonstrates performance gains of AutoMix for various classification tasks on \textbf{eight classification benchmarks}, including CIFAR-10/100~\cite{krizhevsky2009learning}, Tiny-ImageNet~\cite{2017tinyimagenet}, ImageNet-1k~\cite{russakovsky2015imagenet}, CUB-200-2011 (CUB)~\cite{wah2011caltech}, FGVC-Aircraft (Aircraft)~\cite{maji2013fine}, iNaturalist2017/2018 (iNat2017/2018)~\cite{cvpr2018inaturalist}, and Places205~\cite{nips2014place205}. 
We verify generalizabilities of AutoMix for \textbf{eight network architectures}, the experiments adopt popular ConvNets, including ResNet (R)~\cite{he2016deep}, Wide-ResNet (WRN)~\cite{bmvc2016wrn}, ResNeXt (32x4d) (RX)~\cite{xie2017aggregated}, MobileNet.V2~\cite{cvpr2018mobilenetv2}, EfficientNet~\cite{icml2019efficientnet}, and ConvNeXt~\cite{2022convnet}, and Transformer-based architectures (DeiT~\cite{icml2021deit} and Swin Transformer (Swin)~\cite{iccv2021swin}) as backbone networks. 
For a fair comparison, we use the open-source codebase OpenMixup~\cite{2022openmixup} for most mixup methods: 
(i) \textit{hand-crafted} methods: Mixup~\cite{zhang2017mixup}, CutMix~\cite{yun2019cutmix}, ManifoldMix~\cite{verma2019manifold}, AugMix~\cite{hendrycks2019augmix}, AttentiveMix~\cite{icassp2020attentive}, SaliencyMix~\cite{uddin2020saliencymix}, FMix~\cite{harris2020fmix}, and ResizeMix~\cite{qin2020resizemix}; 
(ii) \textit{optimization-based} methods: PuzzleMix~\cite{kim2020puzzle}, Co-Mixup~\cite{kim2021comixup}, and SuperMix~\cite{dabouei2021supermix}. 
Notice that AugMix is reproduced by timm~\cite{wightman2021rsb}, $*$ denotes open-source arXiv preprint work, and $\dag$ denotes the results reproduced by the official source code (Co-Mixup, AlignMix~\cite{2021alignmix}, and TransMix~\cite{2021transmix}). 
All mixup augmentation methods use the optimal $\alpha$ among $\{0.2, 0.5, 1, 2, 4\}$, while the rest of the hyper-parameters follow the original paper. 
AutoMix uses the same set of hyper-parameters in all experiments: $\alpha=2$, the feature layer $l=3$, and the momentum coefficient in MP starts from $m=0.999$ and is increased to $1$ in a cosine curve. 
As for all classification results, we report the \textit{mean} performance of 3 trials where the \textit{median} of top-1 test accuracy in the last 10 training epochs is recorded for each trial, and \textbf{bold} and \blue{blue} denote the best and second best results.

\vspace{-14pt}
\begin{table}[H]
    \centering
    \caption{Top-1 accuracy (\%)$\uparrow$ of various algorithms based on ResNet variants for small-scale classification on CIFAR-10/100 and Tiny-ImageNet datasets.}
    \vspace{-4pt}
\resizebox{0.85\linewidth}{!}{
    \begin{tabular}{l|cc|ccc|cc}
    \toprule
                    & \multicolumn{2}{c|}{CIFAR-10} & \multicolumn{3}{c|}{CIFAR-100}               & \multicolumn{2}{c}{Tiny-ImageNet} \\
    Method          & R-18          & RX-50         & R-18          & RX-50         & WRN-28-8     & R-18          & RX-50           \\ \hline
    Vanilla         & 95.50         & 96.23         & 78.04         & 81.09         & 81.63        & 61.68         & 65.04           \\
    MixUp           & 96.62         & 97.30         & 79.12         & 82.10         & 82.82        & 63.86         & 66.36           \\
    CutMix          & 96.68         & 97.01         & 78.17         & 81.67         & 84.45        & 65.53         & 66.47           \\
    ManifoldMix     & 96.71         & \blue{97.33}  & 80.35         & \blue{82.88}  & 83.24        & 64.15         & 67.30           \\
    SaliencyMix     & 96.53         & 97.18         & 79.12         & 81.53         & 84.35        & 64.60         & 66.55           \\
    FMix$^*$        & 96.58         & 96.76         & 79.69         & 81.90         & 84.21        & 63.47         & 65.08           \\
    PuzzleMix       & 97.10         & 97.27         & 81.13         & 82.85         & 85.02        & 65.81         & 67.83           \\
    Co-Mixup$^\dag$ & \blue{97.15}  & 97.32         & \blue{81.17}  & 82.91         & \blue{85.05} & \blue{65.92}  & \blue{68.02}    \\
    ResizeMix$^*$   & 96.76         & 97.21         & 80.01         & 81.82         & 84.87        & 63.74         & 65.87           \\
    \bf{AutoMix}    & \bf{97.34}    & \bf{97.65}    & \bf{82.04}    & \bf{83.64}    & \bf{85.18}   & \bf{67.33}    & \bf{70.72}      \\ \hline
    Gain            & \gbf{+0.19}   & \gbf{+0.32}   & \gbf{+0.87}   & \gbf{+0.76}   & \gbf{+0.13}  & \gbf{+1.41}   & \gbf{+2.70}     \\
    \bottomrule
    \end{tabular}
    }
    \label{tab:cls_cifar_tiny}
\end{table}
\vspace{-30pt}

\subsubsection{Small-scale Datasets}
\label{exp:cifar}
\paragraph{\textbf{Settings.}} 
On CIFAR-10/100, \texttt{RandomFlip} and \texttt{RandomCrop} with 4 pixels padding for 32$\times$32 resolutions are basic data augmentations, and we use the following training settings: SGD optimizer with SGD weight decay of 0.0001, the momentum of 0.9, the batch size of 100, and training 800 epochs; the basic learning rate is 0.1 adjusted by Cosine Scheduler~\cite{loshchilov2016sgdr}. 
On Tiny-ImageNet, the basic augmentations include \texttt{RandomFlip} and \texttt{RandomResizedCrop} for 64$\times$64 resolutions, and we use the similar training ingredients as CIFAR except for the basic learning rate of 0.2 and training 400 epochs. 
CIFAR version of ResNet variants~\cite{he2016deep} are used, \textit{i.e.}, replacing the $7\times 7$ convolution and MaxPooling by a $3\times 3$ convolution.

\vspace{-8pt}
\paragraph{\textbf{Classification.}} 
Table~\ref{tab:cls_cifar_tiny} shows small-scale classification results on CIFAR-10/100 and Tiny datasets. Compared to the previous state-of-the-art methods, AutoMix consistently surpasses ManifoldMix (+0.32$\sim$1.94\%), PuzzleMix (+0.16$\sim$0.91\%), and Co-Mixup (+0.13$\sim$0.87\%) based on various ResNet architectures on CIFAR-10/100. Moreover, AutoMix noticeably outperforms previous best algorithms by 1.41\% and 2.70\% on the more challenging Tiny-ImageNet.

\begin{figure}[b]
    \centering
    \vspace{-1.0em}
    \includegraphics[width=1.0\linewidth]{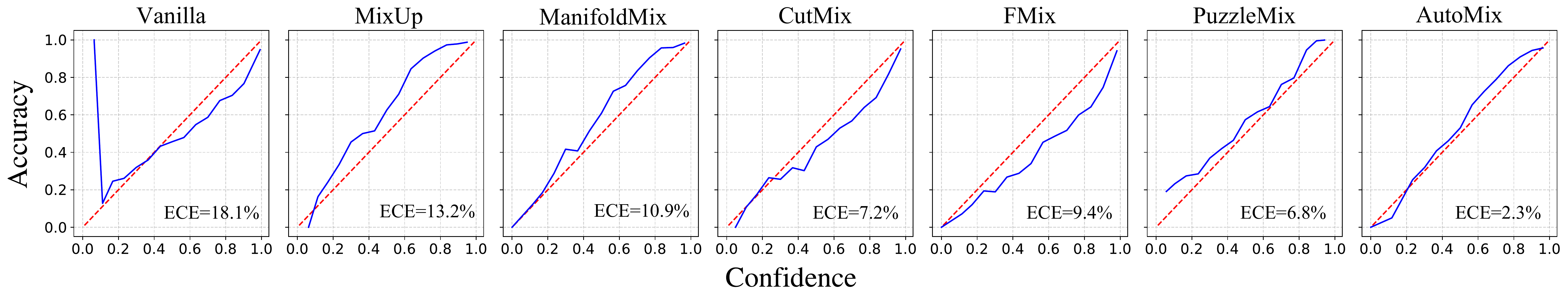}
    \vspace{-2.25em}
    \caption{\textit{Calibration} plots of Mixup variants and AutoMix on CIFAR-100 using ResNet-18. The \textcolor{red}{red} line indicates the expected prediction tendency.}
    \label{fig:calibration}
\end{figure}

\vspace{-8pt}
\paragraph{\textbf{Calibration.}}
DNNs tend to predict over-confidently in classification tasks~\cite{thulasidasan2019mixup}, mixup methods can significantly alleviate this problem. 
To verify the calibration ability of AutoMix, we evaluate popular mixup algorithms by the expected calibration error (ECE) \cite{guo2017calibration} on CIFAR-100, \textit{i.e.}, the absolute discrepancy between accuracy and confidence. 
As shown in Figure \ref{fig:calibration}, AutoMix has the best calibration effect among all competitors, with the ECE error rate of 2.3\%, closest to the red diagonal. 
We can see from the figure that the Cut series does not perform well on calibration, but may further aggravate the overconfidence; while MixUp and ManifoldMix calibrate the predictions, but cause the under-confidence problem.

\begin{table}[H]
    \centering
    \vspace{-1.5em}
    \caption{Top-1 accuracy (\%)$\uparrow$ of image classification based on ResNet variants on ImageNet-1k using PyTorch-style 100-epoch and 300-epoch training procedures.}
    \vspace{-2pt}
\resizebox{0.9\linewidth}{!}{
    \begin{tabular}{l|ccccc|cccc}
    \toprule
                  & \multicolumn{5}{c|}{PyTorch 100 epochs}                                  & \multicolumn{4}{c}{PyTorch 300 epochs}                    \\
    Methods       & R-18         & R-34         & R-50         & R-101        & RX-101       & R-18         & R-34         & R-50         & R-101        \\ \hline
    Vanilla       & 70.04        & 73.85        & 76.83        & 78.18        & 78.71        & \blue{71.83} & 75.29        & 77.35        & 78.91        \\
    MixUp         & 69.98        & 73.97        & 77.12        & 78.97        & 79.98        & 71.72        & 75.73        & 78.44        & 80.60        \\
    CutMix        & 68.95        & 73.58        & 77.17        & 78.96        & 80.42        & 71.01        & 75.16        & 78.69        & 80.59        \\
    ManifoldMix   & 69.98        & 73.98        & 77.01        & 79.02        & 79.93        & 71.73        & 75.44        & 78.21        & 80.64        \\
    SaliencyMix   & 69.16        & 73.56        & 77.14        & 79.32        & 80.27        & 70.21        & 75.01        & 78.46        & 80.45        \\
    FMix$^*$      & 69.96        & 74.08        & 77.19        & 79.09        & 80.06        & 70.30        & 75.12        & 78.51        & 80.20        \\
    PuzzleMix     & \blue{70.12} & \blue{74.26} & \blue{77.54} & \blue{79.43} & 80.53        & 71.64        & \blue{75.84} & 78.86        & \blue{80.67} \\
    ResizeMix$^*$ & 69.50        & 73.88        & 77.42        & 79.27        & \blue{80.55} & 71.32        & 75.64        & \blue{78.91} & 80.52        \\
    \bf{AutoMix}  & \bf{70.50}   & \bf{74.52}   & \bf{77.91}   & \bf{79.87}   & \bf{80.89}   & \bf{72.05}   & \bf{76.10}   & \bf{79.25}   & \bf{80.98}   \\ \hline
    Gain          & \gbf{+0.38}  & \gbf{+0.26}  & \gbf{+0.37}  & \gbf{+0.44}  & \gbf{+0.34}  & \gbf{+0.22}  & \gbf{+0.26}  & \gbf{+0.34}  & \gbf{+0.31}  \\
    \bottomrule
    \end{tabular}
    }
    \vspace{-1.0em}
    \label{tab:cls_in_torch}
\end{table}

\subsubsection{ImageNet Datasets}
\vspace{-0.25em}
\label{exp:imagenet}
\paragraph{\textbf{Settings.}}
In the more challenging large-scale classification scenarios, mixup methods are widely used, especially for recently proposed Transformer-based networks. We evaluate AutoMix and popular mixup variants on ImageNet-1k using three popular training procedures:
(a) PyTorch-style setting trains 100 or 300 epochs by SGD optimizer with the batch size of 256, the basic learning rate of 0.1, the SGD weight decay of 0.0001, and the SGD momentum of 0.9, which is the standard benchmarks for mixup methods~\cite{yun2019cutmix,qin2020resizemix}; 
(b) DeiT setting trains 300 epochs by AdamW optimizer~\cite{iclr2019AdamW} with the batch size of 1024, the basic learning rate of 0.001, and the weight decay of 0.05; 
(c) timm~\cite{wightman2021rsb} RSB A2/A3 settings train 300/100 epochs by LAMB optimizer~\cite{iclr2020lamb} with the batch size of 2048, the basic learning rate of 0.005/0.008, and the weight decay of 0.02. More detailed ingredients and hyper-parameters are provided in Appendix. 
These three settings adopt the basic data augmentations (\texttt{RandomResizedCrop} and \texttt{RandomFlip}) for $224\times 224$ resolutions with Cosine Scheduler by default, (b) and (c) use RandAugment~\cite{cubuk2020randaugment} for better performances.

\begin{figure*}[t]
\vspace{-2.0em}
\begin{minipage}{0.54\linewidth}
\centering
\begin{table}[H]
    \caption{Top-1 accuracy (\%)$\uparrow$ on ImageNet-1k based on various ConvNets using RSB A2/A3 training settings.}
    \vspace{-4pt}
\resizebox{\linewidth}{!}{
    \begin{tabular}{l|c|cc|cc}
    \toprule
                  & R-50         & \multicolumn{2}{c|}{EfficientNet B0} & \multicolumn{2}{c}{MobileNet.V2} \\
    Methods       & A3           & A2                & A3               & A2              & A3             \\ \hline
    RSB           & \blue{78.08} & 77.26             & 74.02            & \blue{72.87}    & 69.86          \\
    MixUp         & 77.66        & 77.19             & 73.87            & 72.78           & \blue{70.17}   \\
    CutMix        & 77.62        & 77.24             & 73.46            & 72.23           & 69.62          \\
    ManifoldMix   & 77.78        & 77.22             & 73.83            & 72.34           & 70.05          \\
    SaliencyMix   & 77.93        & 77.67             & 73.42            & 72.07           & 69.69          \\
    FMix$^*$      & 77.76        & 77.33             & 73.71            & 72.79           & 70.10          \\
    PuzzleMix     & 78.02        & \blue{77.35}      & \blue{74.10}     & 72.85           & 70.04          \\
    ResizeMix$^*$ & 77.85        & 77.27             & 73.67            & 72.50           & 69.94          \\
    \bf{AutoMix}  & \bf{78.44}   & \bf{77.58}        & \bf{74.61}       & \bf{73.19}      & \bf{71.16}     \\ \hline
    Gain          & \gbf{+0.36}  & \gbf{+0.23}       & \gbf{+0.51}      & \gbf{+0.32}     & \gbf{+0.99}    \\
    \bottomrule
    \end{tabular}
    }
    \label{tab:cls_in_rsb}
\end{table}

\end{minipage}
\begin{minipage}{0.45\linewidth}
\centering
\begin{table}[H]
    \caption{Top-1 accuracy (\%)$\uparrow$ on ImageNet-1k based on ViTs and ConvNeXt using DeiT training settings.}
    \vspace{-4pt}
\resizebox{\linewidth}{!}{
    \begin{tabular}{l|ccc}
    \toprule
    Methods         & DeiT-S       & Swin-T       & ConvNeXt-T   \\ \hline
    DeiT            & 79.80        & 81.28        & \blue{82.10} \\
    MixUp           & 79.65        & 81.01        & 80.88        \\
    CutMix          & 79.78        & 81.20        & 81.57        \\
    AttentiveMix    & 77.63        & 77.27        & 78.19        \\
    SaliencyMix     & 79.88        & 81.37        & 81.33        \\
    FMix$^*$        & 77.37        & 79.60        & 81.04        \\
    PuzzleMix       & 80.45        & \blue{81.47} & 81.48        \\
    ResizeMix$^*$   & 78.61        & 81.36        & 81.64        \\
    TransMix$^\dag$ & \blue{80.70} & \bf{81.80}   & -            \\
    \bf{AutoMix}    & \bf{80.78}   & \bf{81.80}   & \bf{82.28}   \\ \hline
    Gain            & \gbf{+0.08}  & \gray{+0.00} & \gbf{+0.18}  \\
    \bottomrule
    \end{tabular}
    }
    \label{tab:cls_in_vits}
\end{table}

\end{minipage}
\vspace{-1.5em}
\end{figure*}

\vspace{-8pt}
\paragraph{\textbf{Classification.}}
Table~\ref{tab:cls_in_torch} and Figure~\ref{fig:intro} show regular image classification results using \textit{only one mixup methods}: AutoMix consistently outperforms previous state-of-the-art methods with light/median/heavy ResNet architectures, \textit{e.g.}, +0.26$\sim$0.44\% for 100 epochs and +0.22$\sim$0.34\% for 300 epochs. 
Table~\ref{tab:cls_in_rsb} and Table~\ref{tab:cls_in_vits} report results on more practical training settings: RSB and DeiT denote \textit{randomly combining Mixup and CutMix} which produces competitive performs as previous state-of-the-art methods (\textit{e.g.}, PuzzleMix), while AutoMix still brings significantly gains over the original RSB (+0.32$\sim$1.30\%) and DeiT (+0.18$\sim$0.98\%). It is worth noticing that previous mixup variants yield little performance gain when adopted on lightweight ConvNets, while AutoMix achieves stable performance gains on these backbones. For example, AutoMix and previous methods improve Vanilla by +0.38\% \textit{vs} 0.08\% based on ResNet-18 in Table~\ref{tab:cls_in_torch}, and improve Vanilla by 0.32\% \textit{vs} -0.08\% based on EfficientNet B0 in Table~\ref{tab:cls_in_rsb}).
Moreover, AutoMix brings remarkable gains over the DeiT setting (0.12$\sim$0.98\%) based on Transformer architectures. AutoMix also yields more competitive performances than the recently proposed Transformer-based mixup method, TransMix.

\subsection{Evaluation on Fine-grained and Scene Classification}
\label{exp:cub200}
\vspace{-2pt}
\paragraph{\textbf{Small-scale datasets.}}
We first perform small-scale fine-grained classification following transfer learning settings on CUB-200 and Aircraft: training 200 epochs by SGD optimizer with the initial learning rate of 0.001, the weight decay of 0.0005, the batch size of 16, using the standard augmentations as in Sec.~\ref{exp:imagenet}; the official PyTorch pre-trained models on ImageNet-1k are adopted as initialization. 
Table~\ref{tab:cls_fgvc_place} shows that AutoMix achieves the best performance and noticeably improves Vanilla (2.19\%/3.55\% on CUB-200 and 1.14\%/1.62\% on Aircraft), which verifies that AutoMix has strong adaptability to more challenging scenarios. 
Since some specific attributes are more useful to distinguish similar classes in fine-grained scenarios, AutoMix generates mixed samples with discriminative patches (\textit{e.g.}, head and beak of birds) rather than a complete object.

\vspace{-8pt}
\paragraph{\textbf{Large-scale datasets.}}
Then, we adopt similar settings as (a) in Sec.~\ref{exp:imagenet} with the total epoch of 100 epochs (training from scratch) on large-scale datasets based on ResNet variants. 
As for the imbalanced and long-tail fine-grained recognition tasks on iNat2017/2018,
Table~\ref{tab:cls_fgvc_place} shows that AutoMix surpasses the previous best methods and improves Vanilla by large margins (2.74\%/4.33\% on iNat2017 and 2.20\%/3.55\% on iNat2018), which demonstrates that AutoMix can alleviate the long-tail and imbalance issues. 
As for scenic classification on Places205, AutoMix still sets state-of-the-art performances. Therefore, we can conclude that AutoMix can adapt to more challenging scenarios.

\begin{table}[t]
    \vspace{-0.5em}
    \centering
    \caption{Top-1 accuracy (\%)$\uparrow$ of various algorithms based on ResNet variants on fine-grained and scenic classification datasets.}
    \vspace{-4pt}
\resizebox{0.99\linewidth}{!}{
    \begin{tabular}{l|cc|cc|cc|cc|cc}
    \toprule
                  & \multicolumn{2}{c|}{CUB-200} & \multicolumn{2}{c|}{FGVC-Aircraft} & \multicolumn{2}{c|}{iNat2017} & \multicolumn{2}{c|}{iNat2018} & \multicolumn{2}{c}{Places205} \\
    Method        & R-18          & RX-50        & R-18             & RX-50           & R-50          & RX-101        & R-50          & RX-101        & R-18          & R-50         \\ \hline
    Vanilla       & 77.68         & 83.01        & 80.23            & 85.10           & 60.23         & 63.70         & 62.53         & 66.94         & 59.63         & 63.10        \\
    MixUp         & 78.39         & 84.58        & 79.52            & 85.18           & 61.22         & 66.27         & 62.69         & 67.56         & 59.33         & 63.01        \\
    CutMix        & 78.40         & 85.68        & 78.84            & 84.55           & 62.34         & 67.59         & 63.91         & 69.75         & 59.21         & 63.75        \\
    ManifoldMix   & \blue{79.76}  & \blue{86.38} & 80.68            & \blue{86.60}    & 61.47         & 66.08         & 63.46         & 69.30         & 59.46         & 63.23        \\
    SaliencyMix   & 77.95         & 83.29        & 80.02            & 84.31           & 62.51         & 67.20         & 64.27         & 70.01         & 59.50         & 63.33        \\
    FMix$^*$      & 77.28         & 84.06        & 79.36            & 86.23           & 61.90         & 66.64         & 63.71         & 69.46         & 59.51         & 63.63        \\
    PuzzleMix     & 78.63         & 84.51        & \blue{80.76}     & 86.23           & \blue{62.66}  & \blue{67.72}  & \blue{64.36}  & \blue{70.12}  & 59.62         & \blue{63.91} \\
    ResizeMix$^*$ & 78.50         & 84.77        & 78.10            & 84.08           & 62.29         & 66.82         & 64.12         & 69.30         & \blue{59.66}  & 63.88        \\
    \bf{AutoMix}  & \bf{79.87}    & \bf{86.56}   & \bf{81.37}       & \bf{86.72}      & \bf{63.08}    & \bf{68.03}    & \bf{64.73}    & \bf{70.49}    & \bf{59.74}    & \bf{64.06}   \\ \hline
    Gain          & \gbf{+0.11}   & \gbf{+0.18}  & \gbf{+0.61}      & \gbf{+0.12}     & \gbf{+0.42}   & \gbf{+0.31}   & \gbf{+0.37}   & \gbf{+0.37}   & \gbf{+0.08}   & \gbf{+0.15}  \\
    \bottomrule
    \end{tabular}
    }
    \vspace{-12pt} 
    \label{tab:cls_fgvc_place}
\end{table}

\begin{figure*}[t]
\vspace{-1.25em}
\begin{minipage}{0.54\linewidth}
\centering
\begin{table}[H]
\centering
    \caption{Top-1 accuracy (\%)$\uparrow$ and FGSM error (\%)$\downarrow$ on CIFAR-100 based on ResNeXt-50 (32x4d) trained 400 epochs.}
    \vspace{-4pt}
\resizebox{0.99\linewidth}{!}{
\begin{tabular}{l|ccc}
    \toprule
                 & Clean              & Corruption         & FGSM                   \\
                 & Acc(\%)$\uparrow$  & Acc(\%)$\uparrow$  & Error(\%)$\downarrow$  \\ \hline
    Vanilla      & 80.24              & 51.71              & 63.92                  \\
    MixUp        & 82.44              & \blue{58.10}       & \blue{56.60}           \\
    CutMix       & 81.09              & 49.32              & 76.84                  \\
    AugMix       & \gray{81.18}       & \gray{66.54}       & \gray{55.59}           \\
    PuzzleMix    & \blue{82.76}       & 57.82              & 63.71                  \\
    \bf{AutoMix} & \bf{83.13}         & \bf{58.35}         & \bf{55.34}             \\
    \bottomrule
    \end{tabular}
    }
\label{tab:cifar_c}
\end{table}


\end{minipage}
\begin{minipage}{0.45\linewidth}
\centering
\begin{table}[H]
\centering
    \caption{Trasfer learning of object detection task with Faster-RCNN on Pascal VOC and COCO datasets.}
    \vspace{-4pt}
\resizebox{0.99\linewidth}{!}{
\begin{tabular}{l|c|ccc}
    \toprule
                 & VOC         & \multicolumn{3}{c}{COCO}                      \\
    Methods      & mAP         & mAP         & AP$^{bb}_{50}$ & AP$^{bb}_{75}$ \\ \hline
    Vanilla      & 81.0        & 38.1        & 59.1           & 41.8           \\
    Mixup        & 80.7        & 37.9        & 59.0           & 41.7           \\
    CutMix       & 81.9        & 38.2        & 59.3           & 42.0           \\
    PuzzleMix    & 81.9        & 38.3        & 59.3           & 42.1           \\
    ResizeMix    & \blue{82.1} & \blue{38.4} & \blue{59.4}    & \blue{42.1}    \\
    \bf{AutoMix} & \bf{82.4}   & \bf{38.6}   & \bf{59.5}      & \bf{42.2}      \\
    \bottomrule
    \end{tabular}
    }
\label{tab:transfer}
\end{table}

\end{minipage}
\vspace{-1.5em}
\end{figure*}

\vspace{-6pt}
\subsection{Robustness}
\vspace{-2pt}
\label{exp:robust}
We first evaluate robustness against corruptions on CIFAR-100-C~\cite{hendrycks2019benchmarking}, which is designed for evaluating the corruption robustness and provides 19 different corruptions (\textit{e.g.}, noise, blur, and digital corruption, \textit{etc}). 
AugMix~\cite{hendrycks2019augmix} is proposed to improve robustness against natural corruptions by minimizing Jensen-Shannon divergence (JSD) between logits of a clean image and two AugMix images.
However, the improvement of AugMix is very limited to clean data. In Table~\ref{tab:cifar_c}, AutoMix shows a consistent top level in both clean and corruption data. 
We further study robustness against the FGSM~\cite{iclr2015fgsm} white box attack of 8/255 $\ell_{\infty}$ epsilon ball following \cite{zhang2017mixup}, and AutoMix outperforms previous methods in Table~\ref{tab:cifar_c}.

\subsection{Transfer Learning}
\vspace{-2pt}
\label{exp:transfer}
\paragraph{\textbf{Weakly supervised object localization.}} 
Following CutMix, we also evaluate AutoMix on the weakly supervised object localization (WSOL) task on CUB-200, which aims to localize objects of interest without bounding box supervision. 
We use CAM to extract attention maps, and calculate the maximal box accuracy with a threshold $\delta \in \{0.3, 0.5, 0.7\}$, following MaxBoxAccV2~\cite{choe2020evaluating}. 
Table~\ref{tab:cub_wsol} shows that AutoMix achieves the best performance to localize semantic regions.

\vspace{-8pt}
\paragraph{\textbf{Object detection.}}
We then evaluate transferable abilities of the learned features to object detection task with Faster R-CNN~\cite{ren2015faster} on PASCAL VOC~\textit{trainval07+12}~\cite{2010pascalvoc} and COCO~\textit{train2017}~\cite{eccv2014MSCOCO} based on Detectron2~\cite{wu2019detectron2}. We fine-tune Faster R-CNN with R50-C4 pre-trained on ImageNet-1k with mixup methods on VOC (24k iterations) and COCO (2$\times$ schedule). Table~\ref{tab:transfer} shows that AutoMix achieves better performances than previous cutting-based mixup variants.

\begin{table}[H]
\centering
    \vspace{-1.5em}
    \caption{MaxBoxAcc (\%)$\uparrow$ for the WSOL task on CUB-200 based on ResNet variants.}
    \vspace{-4pt}
\resizebox{0.80\linewidth}{!}{
\begin{tabular}{lccccccc}
    \toprule
    Backbone & Vanilla & Mixup & CutMix & FMix$^*$     & PuzzleMix & Co-Mixup     & \bf{Ours}  \\ \hline
    R-18     & 49.91   & 48.62 & 51.85  & 50.30        & 53.95     & \blue{54.13} & \bf{54.46} \\
    RX-50    & 53.38   & 50.27 & 57.16  & \blue{59.80} & 59.34     & 59.76        & \bf{61.05} \\
    \bottomrule
    \end{tabular}
    }
    \vspace{-2.5em} 
    \label{tab:cub_wsol}
\end{table}

\begin{figure*}[t]
\vspace{-2.0em}
\begin{minipage}{0.29\linewidth}
\centering
\begin{table}[H]
\centering
    \caption{Ablation of modules in MixBlock.}
    \vspace{-2pt}
\resizebox{\linewidth}{!}{
    \begin{tabular}{l|cc}
    \toprule
                         & \multicolumn{2}{c}{Tiny-ImageNet} \\
    module               & R-18            & RX-50           \\ \hline
    (random grids)       & 64.40           & 66.83           \\
    +cross attention     & 66.87           & 69.76           \\
    +$\lambda$ embedding & 67.15           & 70.41           \\
    +$\ell_{\lambda}$    & \bf{67.33}      & \bf{70.72}      \\
    \toprule
    \end{tabular}
}
\label{tab:ablation_mb}
\end{table}


\end{minipage}
\begin{minipage}{0.70\linewidth}
\centering
    \begin{table}[H]
\centering
    \caption{Ablation of the proposed momentum pipeline (MP) and the cross-entropy loss $l_{CE}$ (CE) based on ResNet-18.}
    \vspace{-3pt}
\resizebox{\linewidth}{!}{
    \begin{tabular}{l|ccc|ccc|ccc}
    \toprule
             & \multicolumn{3}{c|}{CIFAR-100}                 & \multicolumn{3}{c|}{Tiny-ImageNet}             & \multicolumn{3}{c}{ImageNet-1k}                \\
    modules  & MixUp      & CutMix     & $\mathcal{M}_{\phi}$ & MixUp      & CutMix     & $\mathcal{M}_{\phi}$ & MixUp      & CutMix     & $\mathcal{M}_{\phi}$ \\ \hline
    (none)   & 79.12      & 78.17      & 79.46                & 63.39      & 64.40      & 64.84                & 69.98      & 68.95      & 70.04                \\
    +MP(m=0) & -          & -          & 81.75                & -          & -          & 67.05                & -          & -          & 70.41                \\
    +MP      & \bf{80.82} & 79.57      & 81.93                & 66.02      & \bf{65.72} & 67.19                & \bf{70.13} & 70.02      & 70.45                \\
    +MP+CE   & 80.41      & \bf{79.64} & \bf{82.04}           & \bf{66.10} & 65.05      & \bf{67.33}           & 70.10      & \bf{70.04} & \bf{70.50}           \\
    \toprule
    \end{tabular}
}
\label{tab:ablation_mb_mp}
\end{table}

\end{minipage}
\vspace{-1.5em}
\end{figure*}

\subsection{Ablation Study}
\label{exp:ablation}
We conduct an ablation study to prove that each component of AutoMix plays an essential role to make the framework operate properly. 
Three main questions are answered here: 
(1) Are the modules in MB effective?
(2) How many gains can MB bring without EMA and CE? 
(3) Is AutoMix robust to hyperparameters?

\begin{figure*}[t]
\vspace{-2.0em}
\centering
\begin{minipage}{0.52\linewidth}
    \caption{Ablation of hyperparameter $\alpha$ of AutoMix on CIFAR-100 and Tiny-ImageNet.}
    \includegraphics[width=1.0\linewidth]{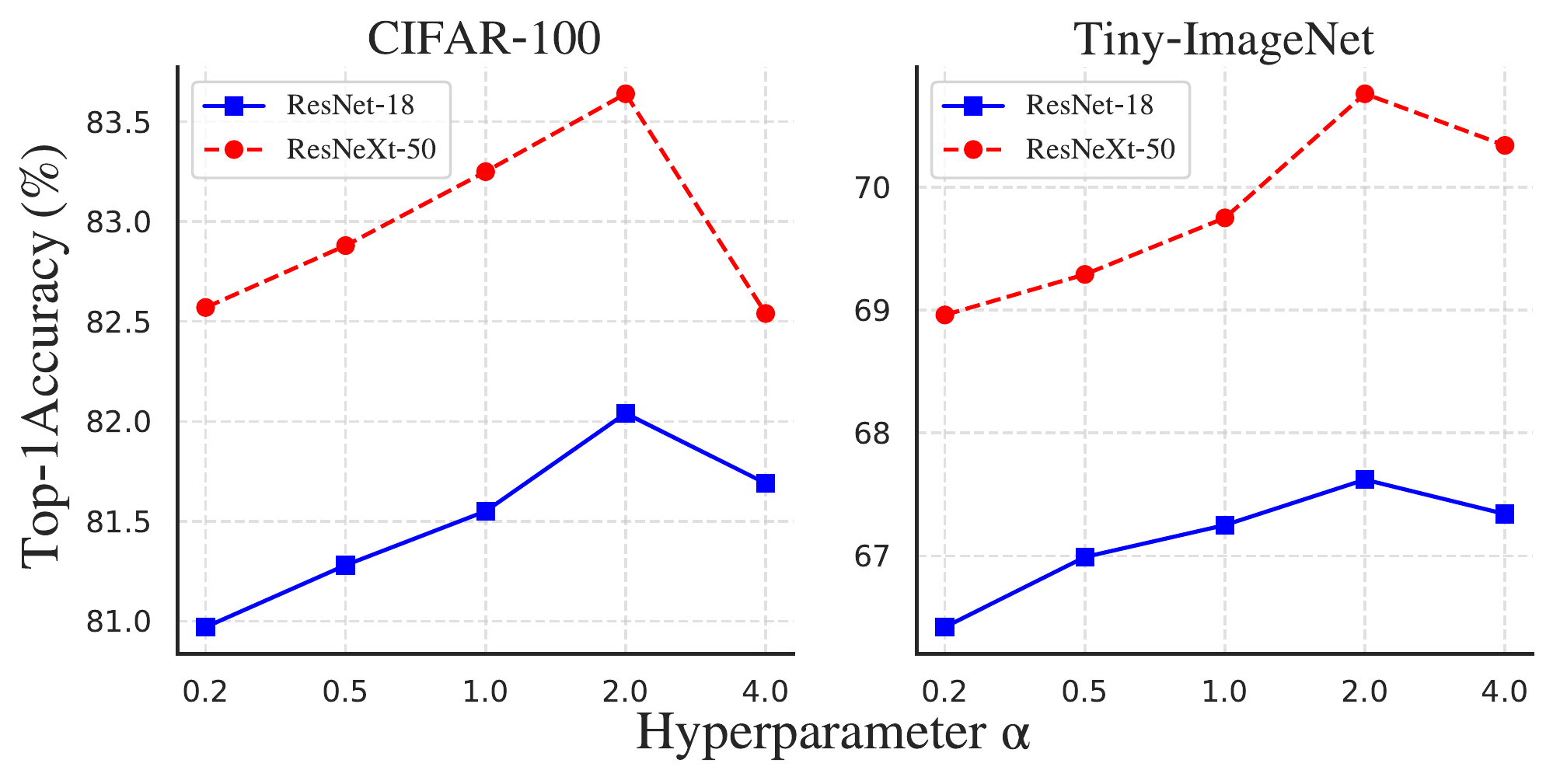}
    \vspace{-25pt}
    \label{fig:alpha}
\end{minipage}
\begin{minipage}{0.42\linewidth}
\begin{table}[H]
\centering
    \caption{Ablation of feature layer $l$ on Tiny-ImageNet, reporting top-1 Acc (\%)$\uparrow$ \textit{vs.} params (M)$\downarrow$ \textit{vs.} the total training time (hours)$\downarrow$.}
    \vspace{-2pt}
\resizebox{\linewidth}{!}{
    \begin{tabular}{c|ccc|ccc}
    \toprule
                        & \multicolumn{3}{c|}{R-18}               & \multicolumn{3}{c}{RX-50} \\
                        & Acc(\%)      & Params       & Time      & Acc(\%)      & Params       & Time       \\ \hline
                  Mixup & \gray{63.86} & \gray{11.27} & \gray{20} & \gray{66.36} & \gray{23.38} & \gray{113} \\
                  $l_1$ & 67.30        & 11.38        & 67        & 70.70        & 23.80        & 413      \\
                  $l_2$ & 67.27        & 11.39        & 41        & 70.43        & 23.86        & 252      \\
\rowcolor{gray94} $l_3$ & \bf{67.33}   & 11.44        & 34        & \bf{70.72}   & 24.84        & 196      \\
                  $l_4$ & 67.32        & 11.64        & 28        & 70.67        & 27.99        & 174      \\
    \bottomrule
    \end{tabular}
    }
    \label{tab:layer_param}
\end{table}

\end{minipage}
\vspace{-1.5em}
\end{figure*}

\begin{itemize}
    \item[(1)] 
    The cross-attention mechanism enables MB to capture the task-relevant pixels between two samples, which is the core design of MB to generate useful mixed masks. 
    Based on this, $\lambda$ embedding and $\ell_{\lambda}$ encourage MB to learn proportional correspondence on a different scale. 
    Without these modules, the performance drops by almost 4\% (66.83\% vs. 70.72\%), as shown in Figure~\ref{tab:ablation_mb}.
    \item[(2)]
    In Table~\ref{tab:ablation_mb_mp}, we show that the EMA and CE adopted in the MP improve the performance of MB by ensuring training stability, however, CE is not as effective for other mixup methods. 
    Most importantly, without these them, i.e. EMA and CE, we show MB still delivers significant gains (\textit{e.g.} +2.29\% and +2.21\% on CIFAR-100 and Tiny). 
    Note that $m=0$ indicates removing EMA, which means $f_{\theta_k}$ is a copy of $f_{\theta_q}$ with the same weights. 
    Therefore, we can confirm the effectiveness of $\mathcal{M}_\phi$.
    \item[(3)]
    AutoMix has two core hyper-parameters, $\alpha$ and $l$, which are fixed for all experiments. 
    A larger $\alpha$ facilitates MB to learn intra-class relationships.
    Figure~\ref{fig:alpha} shows that AutoMix with $\alpha=2$ as default achieves the best performances on various datasets.
    The feature layer $l_3$ makes a good trade-off between the performance and complexity, as shown in Table~\ref{tab:layer_param}.
\end{itemize}

\vspace{-1.0em}
\section{Related Work}
\label{sec: relatedworks}
MixUp~\cite{zhang2017mixup}, the first mixing-based data augmentation algorithm, was proposed to generate mixed samples with mixed labels by convex interpolations of any two samples and their unique one-hot labels. ManifoldMix~\cite{verma2019manifold} extends MixUp to the hidden space of DNNs and \cite{faramarzi2020patchup,2021alignmix} improves ManifoldMix. 
CutMix~\cite{yun2019cutmix} incorporates the Dropout strategy into the mixup strategy and proposes a mixing strategy based on the patch of the image, \textit{i.e.}, randomly replacing a local rectangular area in images. Based on CutMix, AttentiveMix~\cite{icassp2020attentive} and SaliencyMix~\cite{uddin2020saliencymix} guide mixing patches by saliency regions in the image (based on CAM or a saliency detector) to obtain mixed samples with more class-relevant information; ResizeMix~\cite{qin2020resizemix} maintains the information integrity by replacing one resized image directly into a rectangular area of another image; FMix~\cite{harris2020fmix} transforms images into the spectrum domain to generate binary masks by setting a threshold; other researchers design refined mixing strategies~\cite{Baek2021gridmix,cvpr2021stylemix,cvpr2022pixmix},
Furthermore, PuzzleMix~\cite{kim2020puzzle} and Co-Mixup~\cite{kim2021comixup} propose combinatorial optimization strategies to find optimal mixup masks by maximizing the saliency information. Compared with previous methods, AutoMix does not require a hand-crafted sample mixing strategy or saliency information but adaptively generates mixed samples based on mixing ratios and feature maps in an end-to-end manner.

\section{Conclusion and Limitations}
\label{sec:conclusion}
In this paper, we propose an \textit{AutoMix} framework, which optimizes both the mixed sample generation task and the mixup classification task in a momentum training pipeline. Without adding cost to inference, AutoMix can generate out-of-manifold samples with adaptive masks. Extensive experiments have shown the effectiveness and excellent generalizability of the proposed AutoMix on CIFAR, ImageNet, and fine-grained datasets. On top of that, we also outperformed other mixup algorithms when comparing with robustness and localization tasks as well. Furthermore, the proposed momentum training pipeline serves as a significant improvement in convergence speed and overall performance.

As for future work, we consider improving AutoMix in four aspects.
(\romannumeral1) AutoMix is now learning the mixed policy by the proposed cross-attention-based module between only two samples, and it would be more efficient if it could be extended to multiple samples. (\romannumeral2) Supervised labels are required to learn the online mixup policy in AutoMix, which limits the AutoMix to supervised tasks. It would be a general mixup strategy if we extend AutoMix to task-agnostic visual representation learning. (\romannumeral3) Although the time complexity of AutoMix is faster than that of the combinatorial optimization-based methods, there is still a big gap with the hand-crafted methods. A pre-trained Mix Block will be a promising avenue in future research. (\romannumeral4) Despite mixup augmentation techniques are widely studied and used on classification tasks, mixups applied in various downstream tasks are still limited to some variants of Mixup~\cite{zhang2017mixup} and CutMix~\cite{yun2019cutmix} (\textit{e.g.}, Yolo.V4~\cite{2020YOLOv4} employs Mixup and CutMix for object detection). It would benefit downstream tasks if we can extend AutoMix to object detection and instance segmentation with limited training samples. For example, AutoMix might be used as PuzzleMix~\cite{kim2020puzzle} according to the design of CycleMix~\cite{cvpr2022cyclemix} on medical image segmentation tasks.

\section*{Acknowledgement}
This work is supported by the Science and Technology Innovation 2030- Major Project (No. 2021ZD0150100) and the National Natural Science Foundation of China (No. U21A20427). This work was performed during the internship of Zhiyuan Chen at Westlake University. We thank Jianzhu Guo, Cheng Tan, and all reviewers for polishing the writing.

\clearpage
%
%
\bibliographystyle{splncs04}
\bibliography{automix}

\clearpage
\appendix
\section{Appendix}
\label{app}

\subsection{More Implementation Details}
\label{app:implement}
\paragraph{\textbf{Dataset information.}}
We briefly introduce image datasets used in Section \ref{sec:expt}. 
(1) Small scale classification benchmarks: CIFAR-10/100~\cite{krizhevsky2009learning} contains 50,000 training images and 10,000 test images in 32$\times$32 resolutions, with 10 and 100 classes settings. 
(2) Large scale classification benchmarks: ImageNet-1k (IN-1k)~\cite{krizhevsky2012imagenet} contrains 1,281,167 training images and 50,000 validation images of 1000 classes. 
Tiny-ImageNet (Tiny)~\cite{2017tinyimagenet} is a re-scale version of ImageNet-1k, which has 10,000 training images and 10,000 validation images of 200 classes in 64$\times$64 resolutions. 
(3) Small-scale fine-grained classification scenarios: CUB-200-2011 (CUB)~\cite{wah2011caltech} contains 11,788 images from 200 wild bird species for fine-grained classification. FGVC-Aircraft (Aircraft)~\cite{maji2013fine} contains 10,000 images of 100 classes of aircrafts. 
(4) Large-scale fine-grained classification scenarios: iNaturalist2017 (iNat2017)~\cite{cvpr2018inaturalist} contains a total of 5,089 categories with 579,184 training images and 95,986 validation images. iNaturalist2018 (iNat2018)~\cite{cvpr2018inaturalist} contains a total of 8,142 categories with 437,512 training images and 24,426 validation images.
(5) Scenic classification dataset Places205~\cite{nips2014place205} contains around 2,500,000 images from 205 common scene categories. 
Notice that we use modified structures~\cite{he2016deep} of ResNet and ResNeXt for CIFAR-10/100 and Tiny-ImageNet experiments, \textit{i.e.}, replacing the $7\times 7$ convolution and MaxPooling by a $3\times 3$ convolution, while using normal structures on other datasets.

\vspace{-0.75em}
\paragraph{\textbf{Training settings.}}
Detailed training settings of PyTorch~\cite{nips2019pytorch}, DeiT~\cite{icml2021deit}, and RSB A2/A3~\cite{wightman2021rsb} on ImageNet-1k are provided in Table~\ref{tab:app_train_settings}. Notice that we replace the step learning rate decay by Cosine Scheduler~\cite{loshchilov2016sgdr} and remove \texttt{ColorJitter} and \texttt{PCA lighting} in PyTorch training setting for better performances.

\vspace{-0.75em}
\paragraph{\textbf{Reproduction settings.}}
We adopt OpenMixup\footnote{\url{https://github.com/Westlake-AI/openmixup}} implemented in PyTorch~\cite{nips2019pytorch} as the open-source codebase, where we implement AutoMix and reproduce most comparison methods (Mixup~\cite{zhang2017mixup}, CutMix~\cite{yun2019cutmix}, ManifoldMix~\cite{verma2019manifold}, PuzzleMix~\cite{kim2020puzzle}, SaliencyMix~\cite{uddin2020saliencymix}, FMix~\cite{harris2020fmix}, and ResizeMix~\cite{qin2020resizemix}). 
Notice that \textit{optimization-based} methods adopt a consistent $\alpha$ for all datasets, PuzzleMix adopts $\alpha=1$, Co-Mixup and AutoMix adopts $\alpha=2$. 
\textit{Hand-crafted} methods use dataset-specific hyper-parameter settings as follows: 
For CIFAR-10/100, Mixup and ResizeMix use $\alpha=1$, and CutMix, FMix and SaliencyMix use $\alpha=0.2$, and ManifoldMix uses $\alpha=2$, respectively. 
For Tiny-ImageNet and ImageNet-1k using PyTorch-style training settings, ManifoldMix uses $\alpha=0.2$, the rest methods use $\alpha=0.2$ for ResNet-18 while adopt $\alpha=1$ for median and large backbones (\textit{e.g.}, ResNet-50). For iNat2017 and iNat2018, Mixup and ManifoldMix use $\alpha=0.2$, the rest methods adopt $\alpha=1$ for ResNet-50 and ResNeXt-101 while use $\alpha=0.2$ for ResNet-18.
For ImageNet-1k using DeiT and RSB A2/A3 settings and Places205 using PyTorch-style settings, all these methods use $\alpha=0.2$. 
For small-scale fine-grained datasets (CUB-200 and Aircraft), SaliencyMix and FMix use $\alpha=0.2$, and ManifoldMix uses $\alpha=0.5$, while the rest use $\alpha=1$. 
As for other methods, we reproduce results of AugMix~\cite{hendrycks2019augmix}, Co-Mixup~\cite{kim2021comixup}, and SuperMix~\cite{dabouei2021supermix} with their official implementations.

\begin{table}[H]
\centering
    \vspace{-1.5em}
    \caption{Ingredients and hyper-parameters used for ImageNet-1k training settings.}
    \vspace{-4pt}
\resizebox{0.68\linewidth}{!}{
\begin{tabular}{l|cccc}
	\toprule
	Procedure                  & PyTorch   & DeiT              & RSB A2            & RSB A3            \\ \hline
	Train Res                  & 224       & 224               & 224               & 160               \\
	Test Res                   & 224       & 224               & 224               & 224               \\
	Test  crop ratio           & 0.875     & 0.875             & 0.95              & 0.95              \\
	Epochs                     & 100/300   & 300               & 300               & 100               \\ \hline
	Batch size                 & 256       & 1024              & 2048              & 2048              \\
	Optimizer                  & SGD       & AdamW             & LAMB              & LAMB              \\
	LR                         & 0.1       & $1\times 10^{-3}$ & $5\times 10^{-3}$ & $8\times 10^{-3}$ \\
	LR decay                   & cosine    & cosine            & cosine            & cosine            \\
	Weight decay               & $10^{-4}$ & 0.05              & 0.02              & 0.02              \\
	Warmup epochs              & \xmarkg   & 5                 & 5                 & 5                 \\ \hline
	Label smoothing $\epsilon$ & \xmarkg   & 0.1               & \xmarkg           & \xmarkg           \\
	Dropout                    & \xmarkg   & \xmarkg           & \xmarkg           & \xmarkg           \\
	Stoch. Depth               & \xmarkg   & 0.1               & 0.05              & \xmarkg           \\
	Repeated Aug               & \xmarkg   & \cmark            & \cmark            & \xmarkg           \\
	Gradient Clip.             & \xmarkg   & 1.0               & \xmarkg           & \xmarkg           \\ \hline
	H. flip                    & \cmark    & \cmark            & \cmark            & \cmark            \\
	RRC                        & \cmark    & \cmark            & \cmark            & \cmark            \\
	Rand Augment               & \xmarkg   & 9/0.5             & 7/0.5             & 6/0.5             \\
	Auto Augment               & \xmarkg   & \xmarkg           & \xmarkg           & \xmarkg           \\
	Mixup alpha                & \xmarkg   & 0.8               & 0.1               & 0.1               \\
	Cutmix alpha               & \xmarkg   & 1.0               & 1.0               & 1.0               \\
	Erasing prob.              & \xmarkg   & 0.25              & \xmarkg           & \xmarkg           \\
	ColorJitter                & \xmarkg   & \xmarkg           & \xmarkg           & \xmarkg           \\ \hline
	EMA                        & \xmarkg   & \cmark            & \xmarkg           & \xmarkg           \\ \hline
	CE loss                    & \cmark    & \cmark            & \xmarkg           & \xmarkg           \\
	BCE loss                   & \xmarkg   & \xmarkg           & \cmark            & \cmark            \\ \hline
	Mixed precision            & \xmarkg   & \xmarkg           & \cmark            & \cmark            \\
    \bottomrule
    \end{tabular}
    }
    \label{tab:app_train_settings}
    \vspace{-1.0em}
\end{table}

\subsection{More Experiments and Ablations}
\label{app:exp&ablation}
\paragraph{\textbf{More Experiments.}}
We evaluate AutoMix for various training epochs on CIFAR-10/100 based on ResNet-18 (R-18) and ResNeXt-50 (RX-50), as shown in Table~\ref{tab:app_cifar10_r18_rx50} and Table~\ref{tab:app_cifar100_r18_rx50}. 
It is worth noting that some methods converge fast while suffering performance decay with longer train times, such as CutMix and SaliencyMix, and some methods perform better when train longer, such as ManifoldMix training 1200 epochs.
Unlike these methods, AutoMix steadily outperforms them by a large margin regardless of the training time setting.

\begin{figure}[t]
    \centering
    \includegraphics[width=1.0\linewidth]{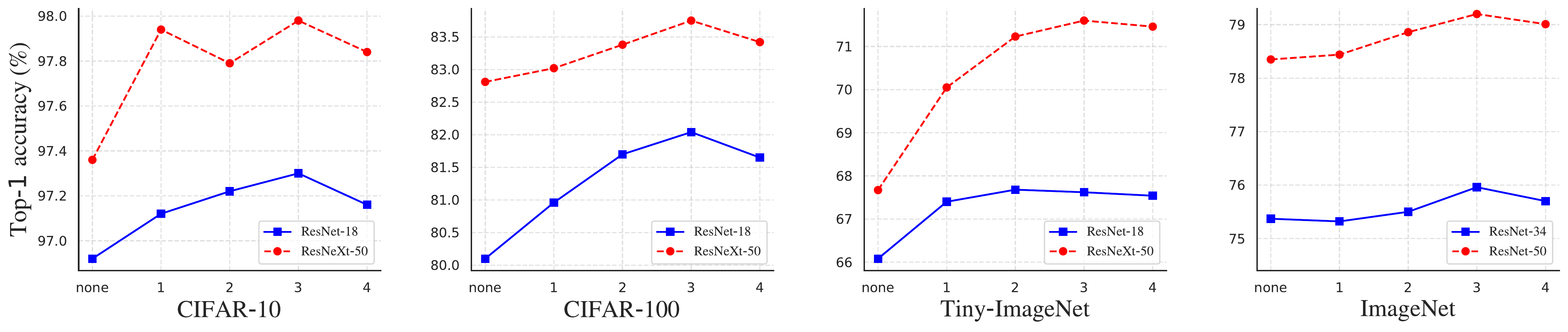}
    \vspace{-2.5em}
    \caption{Top-1 accuracy ablation study on feature layer $l$.}
    \label{fig:app_ablation}
\end{figure}

\vspace{-0.5em}
\paragraph{\textbf{Hyperparameters for AutoMix.}}
We further analyze the hyper-parameter setting for AutoMix with extra ablation studies conducted on Tiny-ImageNet and ImageNet-1k with various network architectures. As the same conclusion we provided in main body of experiment, the result in Figure~\ref{fig:app_ablation} also recommends the choice of $l=3$, which reflects the hyper-parameter robustness of AutoMix.

\begin{table}[t]
\centering
    \caption{Top-1 accuracy (\%)$\uparrow$ on CIFAR-10 based on ResNet-18 and ResNeXt-50 (32x4d) trained with various epochs. $*$ denotes unpublished open-source work on \textit{arxiv}.}
    \vspace{-4pt}
\resizebox{0.85\linewidth}{!}{
    \begin{tabular}{lcccc|cccc}
    \toprule
    Backbone            & \multicolumn{4}{c}{ResNet-18}                             & \multicolumn{4}{c}{ResNeXt-50}                      \\
    Epoch               & 200ep        & 400ep        & 800ep        & 1200ep       & 200ep        & 400ep        & 800ep        & 1200ep \\ \hline
    Vanilla             & 94.87        & 95.10        & 95.50        & 95.59        & 95.92        & 95.81        & 96.23        & 96.26  \\
    MixUp               & 95.70        & 96.55        & 96.62        & 96.84        & 96.88        & 97.19        & 97.30        & 97.33  \\
    CutMix              & 96.11        & 96.13        & 96.68        & 96.56        & 96.78        & 96.54        & 96.60        & 96.35  \\
    ManifoldMix         & 96.04        & 96.57        & 96.71        & 97.02        & 96.97        & 97.39        & \blue{97.33} & \blue{97.36} \\
    SaliencyMix         & 96.05        & 96.42        & 96.20        & 96.18        & 96.65        & 96.89        & 96.70        & 96.60  \\
    FMix$^*$            & 96.17        & 96.53        & 96.18        & 96.01        & 96.72        & 96.76        & 96.76        & 96.10  \\
    PuzzleMix           & \blue{96.42} & 96.87        & \blue{97.10} & 97.03        & \blue{97.05} & 97.24        & 97.27        & 97.34  \\
    ResizeMix$^*$       & 96.16        & \blue{96.91} & 96.76        & \blue{97.04} & 97.02        & \blue{97.38} & 97.21        & 97.36  \\
    \bf{AutoMix}        & \bf{96.59}   & \bf{97.08}   & \bf{97.34}   & \bf{97.30}   & \bf{97.19}   & \bf{97.42}   & \bf{97.65}   & \bf{97.51}  \\ \hline
    Gain                & \gfn{+0.17}  & \gfn{+0.17}  & \gfn{+0.24}  & \gfn{+0.26}  & \gfn{+0.14}  & \gfn{+0.04}  & \gfn{+0.32}  & \gfn{+0.15} \\
    \bottomrule
    \end{tabular}
    }
    \label{tab:app_cifar10_r18_rx50}
\end{table}

\begin{table}[t]
\centering
    \caption{Top-1 accuracy (\%)$\uparrow$ on CIFAR-100 based on ResNet-18 and ResNeXt-50 (32x4d) trained with various epochs.}
    \vspace{-4pt}
\resizebox{0.85\linewidth}{!}{
    \begin{tabular}{lcccc|cccc}
    \toprule
    Backbone            & \multicolumn{4}{c}{ResNet-18}                             & \multicolumn{4}{c}{ResNeXt-50}                      \\
    Epoch               & 200ep        & 400ep        & 800ep        & 1200ep       & 200ep        & 400ep        & 800ep        & 1200ep \\ \hline
    Vanilla             & 76.42        & 77.73        & 78.04        & 78.55        & 79.37        & 80.24        & 81.09        & 81.32  \\
    MixUp               & 78.52        & 79.34        & 79.12        & 79.24        & 81.18        & 82.54        & 82.10        & 81.77  \\
    CutMix              & 79.45        & 79.58        & 78.17        & 78.29        & 81.52        & 78.52        & 78.32        & 77.17  \\
    ManifoldMix         & 79.18        & 80.18        & 80.35        & 80.21        & 81.59        & 82.56        & \blue{82.88} & \blue{83.28} \\
    SaliencyMix         & 79.75        & 79.64        & 79.12        & 77.66        & 80.72        & 78.63        & 78.77        & 77.51  \\
    FMix$^*$            & 78.91        & 79.91        & 79.69        & 79.50        & 79.87        & 78.99        & 79.02        & 78.24  \\
    PuzzleMix           & 79.96        & 80.82        & 81.13        & 81.10        & 81.69        & 82.84        & 82.85        & 82.93  \\
    Co-Mixup            & \blue{80.01} & \blue{80.87} & \blue{81.17} & \blue{81.18} & \blue{81.73} & \blue{82.88} & 82.91        & 82.97  \\
    ResizeMix$^*$       & 79.56        & 79.19        & 80.01        & 79.23        & 79.56        & 79.78        & 80.35        & 79.73  \\
    \bf{AutoMix}        & \bf{80.12}   & \bf{81.78}   & \bf{82.04}   & \bf{81.95}   & \bf{82.84}   & \bf{83.32}   & \bf{83.64}   & \bf{83.80}  \\ \hline
    Gain                & \gfn{+0.11}  & \gfn{+0.91}  & \gfn{+0.87}  & \gfn{+0.77}  & \gfn{+1.11}  & \gfn{+0.44}  & \gfn{+0.76}  & \gfn{+0.52} \\
    \bottomrule
    \end{tabular}
    }
    \vspace{-4pt}
    \label{tab:app_cifar100_r18_rx50}
\end{table}

\begin{wrapfigure}{r}{0.5\textwidth}
    \vspace{-4.5em}
    \centering
    \includegraphics[width=0.80\linewidth]{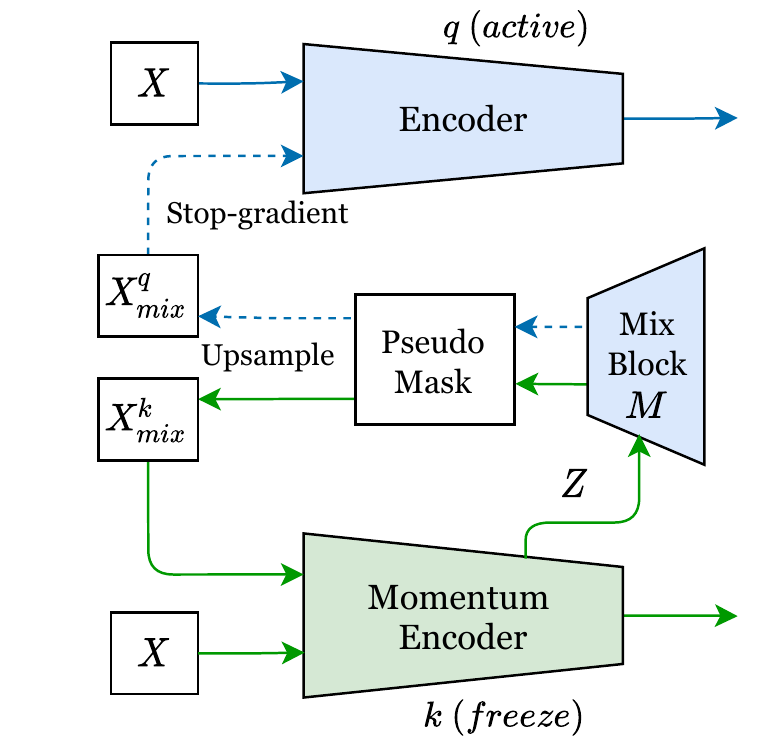}
    \vspace{-1.25em}
    \caption{The network architecture of AutoMix. The parameters in blue modules (active) are updated by backpropagation while the green (freeze) using momentum update in Equation~\ref{eq:momentum}.}
    \label{fig:app_arch}
    \vspace{-2.25em}
\end{wrapfigure}

\subsection{Architecture of Network}
\label{app:network}
The detailed structure of AutoMix is illustrated in Figure~\ref{fig:app_arch}. Similar to the flow chart in the method, the module colored as blue can be updated by backpropagation but not green. Furthermore, the dotted line means stop-gradient. Notice that we use the encoder $k$ for inference and drop $\mathcal{M}_{\phi}$ after training. The training process contains three steps: (1) using the momentum encoder $k$ to generate the feature maps $z$ for $\mathcal{M}_{\phi}$; (2) generating $X_{mix}^q$ and $X_{mix}^k$ based on two mixing factors $\lambda_q$ and $\lambda_k$ and the feature maps; (3) training the active encoder $q$ with mixed samples $X_{mix}^q$ and optimizing $\mathcal{M}_{\phi}$ with $X_{mix}^k$ separately.

\subsection{Algorithm of AutoMix}
\label{app:pseudocode}
We provide the pseudo code of AutoMix in Pytorch style:
\vspace{-10pt}
\begin{algorithm}[H]
\caption{Pseudocode AutoMix in Pytorch style.}
\label{alg:code}
\definecolor{codeblue}{rgb}{0.25,0.5,0.5}
\lstset{
  backgroundcolor=\color{white},
  basicstyle=\fontsize{7.2pt}{7.2pt}\ttfamily\selectfont,
  columns=fullflexible,
  breaklines=true,
  captionpos=b,
  commentstyle=\fontsize{7.2pt}{7.2pt}\color{codeblue},
  keywordstyle=\fontsize{7.2pt}{7.2pt},
}
\begin{lstlisting}[language=python]
# f_q, f_k, M: encoder networks and MixBlock
# lam_q, lam_k: sampled from Beta distribution
# idx_q, idx_k: rearrange index
# m: momentum coefficientt

f_k.params = f_q.params # initialize
for x, y in loader: # load a minibatch

    # two different permutations of data pairs
    x_q, x_k = x[index_q],x[index_k] 
    y_q, y_k = y[index_q],y[index_k]
    lat_f = f_k(x) # hidden representation and logits: NxCxWxH
    
    # generate mixing sample, no gradient to q    
    m_q, m_k = M(x,[lam_q, lam_k],[idx_q, idx_k],lat_f) 
    logits_mix_k = f_k(m_k) # mixed logits: NxC
    logits_cls_q, logits_mix_q = f_q(x),f_q(m_q) # one-hot logits: NxC
    
    # mixup cross-entropy losses for q and M
    loss_cls = ClassificationLoss(lam_q,logits_mix_q,y) # including one-hot CE loss
    loss_gen = GenerationLoss(lam_k,logits_mix_k,y) # including loss_lambda
    loss = loss_cls + loss_gen
    
    loss.backward() 
    update(f_q.params, M.params) # SGD update (q and M)
    f_k_params = m*f_k+(1-m)*f_q.params # momentum update
\end{lstlisting}
\end{algorithm}

\begin{figure*}
    \centering
    \includegraphics[width=0.95\linewidth]{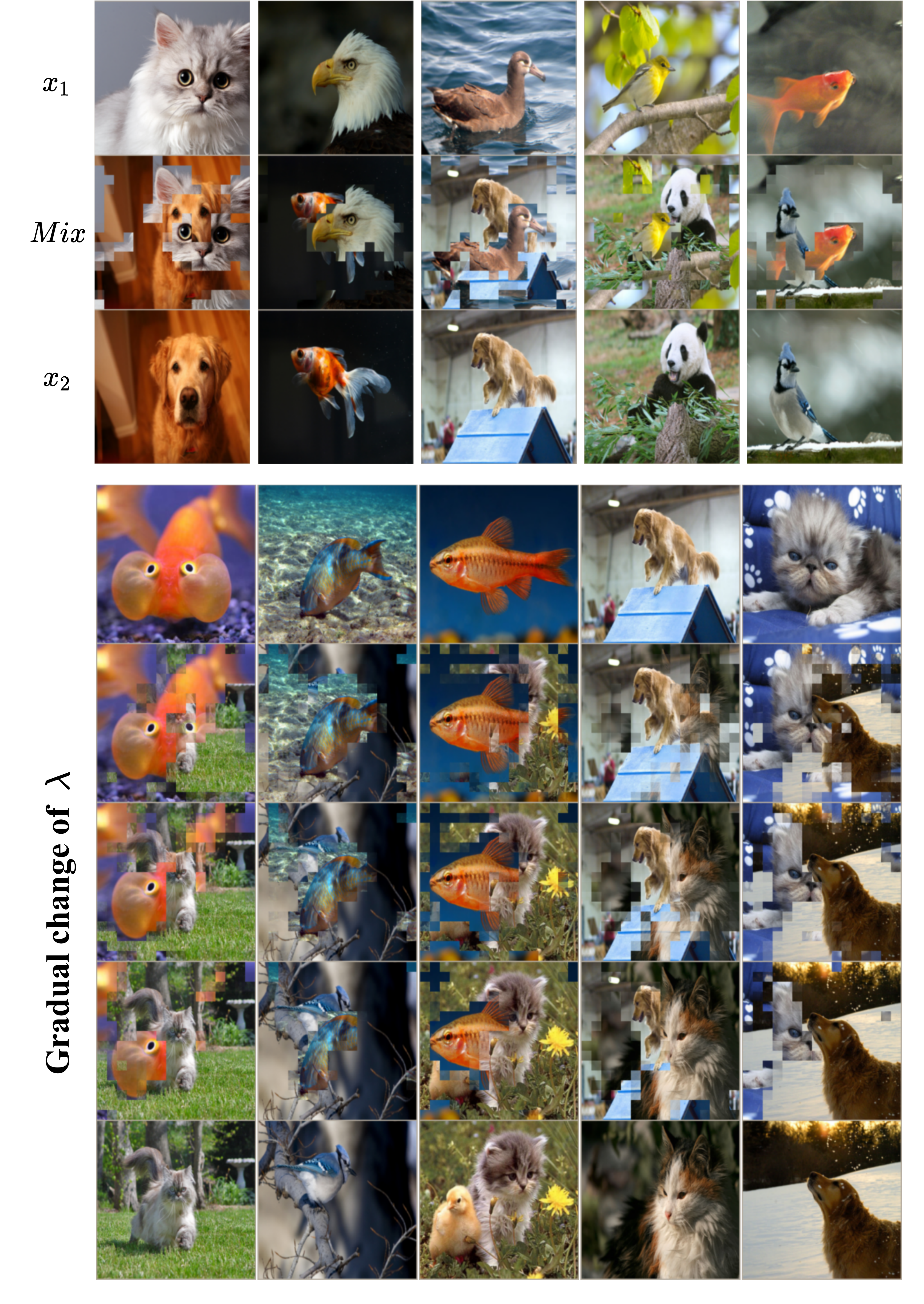}
    \caption{Visualization of mixed samples on ImageNet-1k. The upper part presents the plot of mixed samples from AutoMix ($l=3$) for $\lambda=0.5$; the lower shows the mixed samples when different $\lambda$ values are taken.}
    \label{fig:Visualization_ImageNet_R50.}
\end{figure*}
\begin{figure*}
    \centering
    \includegraphics[width=0.95\linewidth]{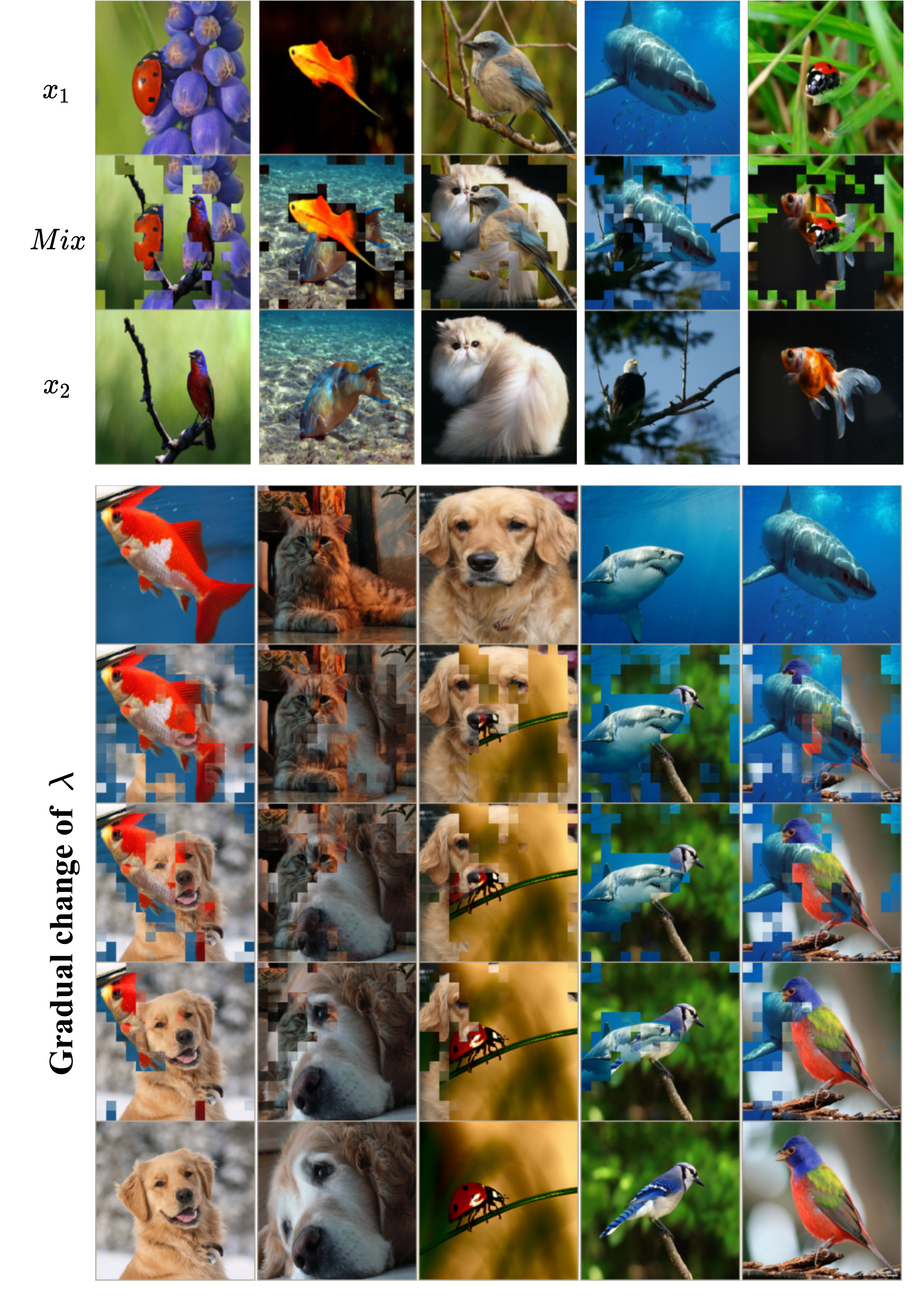}
    \caption{Visualization of mixed samples on ImageNet-1k. The upper part presents the plot of mixed samples from AutoMix ($l=3$) for $\lambda=0.5$; the lower offers the mixed samples when different $\lambda$ values are taken.}
    \label{fig:Visualization_ImageNet_R50_2.}
\end{figure*}

\end{document}